\documentclass{article} 
\usepackage{iclr2019_conference,times}

\usepackage{amsmath,amsfonts,bm}

\def\eqref#1{equation~\ref{#1}}

\def\1{\bm{1}}

\DeclareMathAlphabet{\mathsfit}{\encodingdefault}{\sfdefault}{m}{sl}
\SetMathAlphabet{\mathsfit}{bold}{\encodingdefault}{\sfdefault}{bx}{n}

\usepackage{hyperref}
\usepackage{subfig}
\usepackage{epsfig}
\usepackage{wrapfig}
\usepackage{xspace}
\usepackage{graphicx}
\usepackage{color} 
\usepackage[utf8]{inputenc} 
\usepackage[T1]{fontenc}    
\usepackage{hyperref}       
\usepackage{url}            
\usepackage{booktabs}       
\usepackage{amsfonts}       
\usepackage{nicefrac}       
\usepackage{microtype}      
\usepackage{graphicx}
\usepackage{amsmath}
\usepackage{adjustbox}
\usepackage[colorinlistoftodos,prependcaption,textsize=tiny,disable=True]{todonotes}
\usepackage{wrapfig,lipsum,booktabs}
\newcommand{\myparagraph}[1]{\noindent\textbf{#1}}

\newcommand{\marcus}[1]{\todo[linecolor=green,backgroundcolor=green!10,bordercolor=green]{MR: #1}}

\newcommand{\SNI}{\texttt{SNID}\xspace}
\newcommand{\SNIwoNI}{\texttt{SNI}} 
\newcommand{\SLNIwoNI}{\texttt{SLNI}\xspace}
\newcommand{\SLNI}{\texttt{SLNID}\xspace}
\newcommand{\SNIFull}{Sparse coding through Neural Inhibition (\SNIwoNI)\xspace}
\newcommand{\SLNIFull}{Sparse coding through Local Neural Inhibition (\SLNIwoNI)\xspace}
\newcommand{\SLNIDFull}{Sparse coding through Local Neural Inhibition and Discounting (\SLNI)\xspace}

\def \hfillx {\hspace*{-\textwidth} \hfill}

\title{Selfless Sequential Learning}

\author{Rahaf Aljundi \\
KU Leuven \\
ESAT-PSI, Belgium\\
\texttt{rahaf.aljundi@gmail.com} \\
\And
Marcus Rohrbach  \\
Facebook AI Research\\
\texttt{mrf@fb.com} \\
\AND
Tinne Tuytelaars \\
KU Leuven \\
ESAT-PSI, Belgium \\
\texttt{tinne.tuytelaars@esat.kuleuven.be}
}
\iclrfinalcopy 
\begin{document}

\maketitle

\begin{abstract}
Sequential learning, also called lifelong learning, studies the problem of learning tasks in a sequence with access restricted to only the data of the current task.
In this paper we look at a scenario with fixed model capacity, and postulate that the learning process should not be selfish, i.e. it should account for future tasks to be added and 
thus leave enough capacity for them. 
To achieve \emph{Selfless Sequential Learning} we study 
different regularization strategies and activation functions.
We find that 
imposing sparsity at the level of the representation (i.e.~neuron activations)
is more beneficial for sequential learning than encouraging parameter sparsity.

In particular, we propose a novel regularizer,
that encourages representation sparsity by means of neural inhibition. It results in few active neurons  which in turn leaves more free neurons to be utilized by upcoming tasks. 
As
neural inhibition over an entire layer
can be too drastic, especially for 
complex tasks requiring strong representations,
our regularizer only inhibits other neurons in a local neighbourhood, inspired by lateral inhibition processes in the brain.
We combine our novel regularizer
with  state-of-the-art lifelong learning methods that penalize  changes to important previously learned parts of the network. We show that our new regularizer leads to increased sparsity which translates in consistent performance improvement 
on diverse datasets\footnote{{The code is available at} \mbox{ \url{https://github.com/rahafaljundi/Selfless-Sequential-Learning}}} .

\end{abstract}

\section{Introduction}\label{sec:intro}
Sequential learning, also referred to as continual, incremental, or lifelong learning (LLL), studies the problem of learning a 
sequence of tasks, one at a time,  without access to the training data of previous or future tasks.
When learning a new task, a key challenge in this context is how to 
avoid 
catastrophic interference with 
the tasks learned previously \citep{french1999catastrophic,li2016learning}. 
   Some methods exploit an additional episodic memory to store  a small amount of previous tasks data to regularize future task learning (e.g.~\cite{lopez2017gradient}). Others store 
     previous tasks models and at test time, select one model or merge the models~\citep{rusu2016progressive,aljundi2016expert,lee2017overcoming}. In contrast, in this work we are interested in the challenging situation of learning a sequence of tasks \emph{without access to any previous or future task data} and \emph{restricted to a fixed  model capacity},
     as also studied in \cite{kirkpatrick2016overcoming,aljundi2017memory,fernando2017pathnet,mallya2017packnet,dataovercoming}. 
     This scenario not only has many practical benefits, including privacy and scalability, but also resembles more closely how the mammalian brain learns tasks over time.

The mammalian brain is composed of billions of neurons.
Yet at any given time,
information is represented by only a few active neurons resulting in a sparsity of 90-95\%~\citep{lennie2003cost}.  
In neural biology, \emph{lateral inhibition} describes the 
process where
an activated neuron 
reduces the activity of its weaker neighbors. This creates a powerful decorrelated and compact 
representation with minimum interference between different input patterns in the brain~\citep{yu2014sparse}. 
This is in stark contrast with artificial neural networks, which typically learn dense representations that are highly entangled~\citep{bengio2009learning}. 
Such an entangled representation is quite sensitive to changes in the input patterns, in that it responds differently to input patterns with only small variations. 
\cite{french1999catastrophic} suggests that an overlapped 
representation 
plays a crucial role 
in catastrophic forgetting and reducing this overlap would result in a reduced  interference.
\cite{cogswell2015reducing} show that when the amount of overfitting  in a neural network is reduced, the  representation correlation is also reduced.  As such, learning a disentangled representation is more powerful and less vulnerable to catastrophic interference. However, if the learned  disentangled representation at a given task is  not sparse, only little capacity is left for learning new tasks. 
This would in turn result in  either an underfitting to the new tasks or again a forgetting of previous tasks. In contrast,  a sparse and decorrelated representation would lead to a powerful representation and at the same time enough free neurons that can be changed without interference with the neural activations learned for the previous tasks. 

\begin{figure*}[t]
\centering
\vspace*{-1cm} 
\includegraphics[width=1\textwidth]{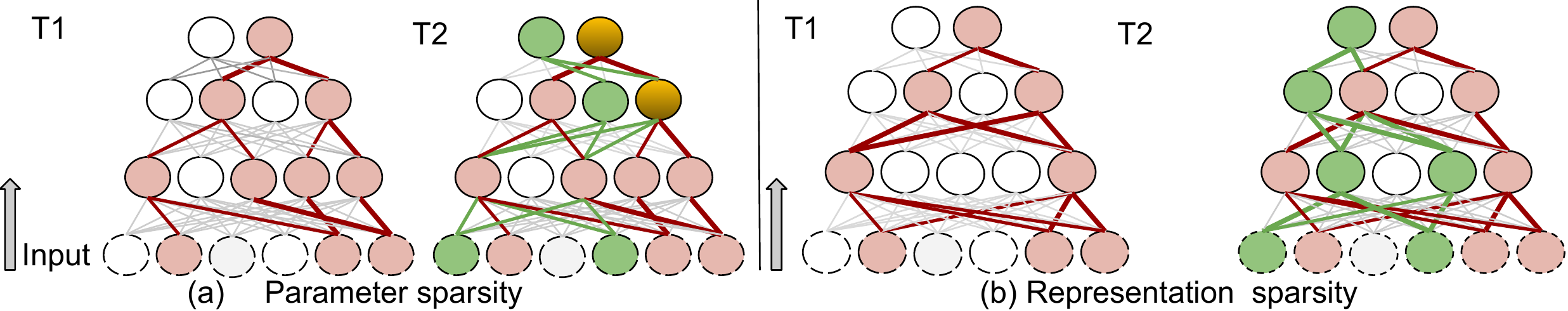} 
  \caption{\footnotesize The difference between parameter sparsity (a) and representation sparsity (b) in a simple two tasks case. First layer indicates input patterns. Learning the first task utilizes parts indicated in red. Task 2 has different input patterns and 
  {uses}
  parts shown in green. Orange indicates changed neurons activations as a result of the second task. 
  In (a), when an example from the first task is encountered again, the activations of the first layer will not be affected by the changes, however, the second and later layer activations are changed.
  Such interference is 
    {largely reduced}
  when
  {imposing}
  sparsity on the representation (b). }  
  \label{fig:param_vs_rep}
 \vspace*{-0.7cm} 
\end{figure*}

In general, sparsity in neural networks can be thought of either in terms of the network parameters or in terms of the representation (i.e., the activations). In this paper we postulate, and confirm 
experimentally, that a sparse and decorrelated
representation is preferable over parameter sparsity 
in a sequential learning scenario. 
There are two arguments for this: first, a
sparse
representation is less sensitive to new and different patterns (such as data from new tasks) and second,  the training procedure of the new tasks can use the free neurons leading to less interference with the previous tasks, hence reducing forgetting.
In contrast, when the effective parameters are spread among different neurons,  changing the ineffective ones would change the function of their corresponding neurons and hence interfere with previous tasks (see also Figure~\ref{fig:param_vs_rep}).
Based on these observations, we  propose a new regularizer that exhibits a behavior similar to the lateral inhibition in  biological neurons. 
The main idea of our regularizer is to penalize 
neurons that are active at the same time. This leads to more sparsity
 and a decorrelated representation. 
However, complex tasks
may actually require multiple active neurons in a layer at the same time to learn a strong representation.  Therefore, our regularizer, \textbf{\SLNIDFull},  only penalizes neurons locally. Furthermore, we don't want inhibition to affect previously learned tasks, even if later tasks use neurons from earlier tasks. An important component of \SLNI is thus to discount inhibition from/to neurons which have high \emph{neuron importance} --  a new concept that we introduce in analogy to parameter importance~\citep{kirkpatrick2016overcoming,zenke2017improved,aljundi2017memory}.
 When combined
 with a state-of-the-art important parameters preservation method \citep{aljundi2017memory,kirkpatrick2016overcoming}, 
 our proposed regularizer leads to  sparse and decorrelated representations which improves the lifelong learning performance.

Our contribution is threefold. First, we direct attention to \textbf{ Selfless Sequential Learning } and study a diverse set of representation based regularizers, parameter based regularizers, as well as 
sparsity inducing activation functions to this end. These 
have not been  studied extensively in the lifelong learning  literature before. Second, we propose a novel regularizer, \SLNI, which is inspired by lateral inhibition in the brain. Third, we show that our proposed regularizer  consistently outperforms alternatives on three diverse datasets (Permuted MNIST, CIFAR, Tiny Imagenet) and we compare to and outperform  state-of-the-art LLL approaches on an  8-task object classification challenge.
 \SLNI can be applied to different regularization based LLL approaches, and we show experiments with MAS \citep{aljundi2017memory} and EWC \citep{kirkpatrick2016overcoming}.

In the following, we first 
discuss 
related LLL approaches 
and 
different regularization criteria from a 
LLL
perspective  (Section~\ref{sec:related}). We proceed by introducing Selfless Sequential Learning and detailing our novel regularizer (Section~\ref{sec:method}). %
Section~\ref{sec:experiments} describes our experimental evaluation, while Section~\ref{sec:conclusion} concludes the paper.

\section{Related Work}\label{sec:related}
The goal in lifelong learning is to learn a sequence of tasks without catastrophic forgetting of previously learned ones~\citep{thrun1995lifelong}. 
One can identify different approaches to introducing lifelong learning in neural networks. Here, 
we focus on learning a sequence of tasks using a fixed model capacity, i.e.~with a fixed architecture and fixed number of parameters. Under this setting, methods either follow a pseudo rehearsal approach,~i.e.~using the new task data to approximate the performance of the previous task~\citep{li2016learning,triki2017encoder}, or aim at identifying the important parameters used by the current set of tasks and penalizing  changes to those parameters by new tasks~\citep{kirkpatrick2016overcoming,zenke2017improved,aljundi2017memory,chaudhry2018riemannian,liu2018rotate}.  To identify the important parameters for a given task, Elastic Weight Consolidation~\citep{kirkpatrick2016overcoming} uses an approximation of the Fisher information matrix computed after training a given task. \cite{liu2018rotate} suggest a network reparameterization to obtain a better diagonal approximation of the Fisher Information matrix of the network  parameters.  Path Integral~\citep{zenke2017improved} estimates the importance of the network parameters while learning a given task by accumulating the contribution of each parameter to the change in the loss. \cite{chaudhry2018riemannian} suggest a KL-divergence based generalization of Elastic Weight Consolidation and Path Integral. Memory Aware Synapses~\citep{aljundi2017memory}  estimates the importance of the parameters in an online manner  without supervision by measuring the sensitivity of the learned function  to small perturbations on the parameters. This method
is less sensitive to the data distribution shift, and 
a local version proposed by the authors resembles applying Hebb rule~\citep{hebb2002organization} to consolidate the important parameters, 
making it more biologically plausible.

A common drawback of all the above methods is that 
learning a task could  utilize a good portion of the network capacity, leaving few "free" neurons to be adapted by the new task. This in turn leads to inferior performance on the newly learned tasks or forgetting the previously learned ones, as we will show in the experiments. Hence, we study the role of sparsity and representation decorrelation in sequential learning.  
This aspect has not received much attention in the literature yet.
Very recently,
\citep{dataovercoming} 
proposed to overcome catastrophic forgetting through learned hard attention masks for each task with L1 regularization imposed on the accumulated hard attention masks. This comes closer to our approach although we  study and propose a regularization scheme on the learned representation.

The concept of reducing the representation overlap has been suggested before in early attempts towards overcoming catastrophic forgetting in neural networks~\citep{french1999catastrophic}. This has led to several methods with the goal of orthogonalizing  the activations~\citep{french1992semi,french1994dynamically,kruschke1992alcove,kruschke1993human,sloman1992reducing}. However, these approaches are mainly designed for specific architectures and activation functions, which makes it hard to integrate them in recent neural network structures. 

The sparsification of neural networks has mostly been studied for compression. SVD decomposition can be applied 
to reduce the number of effective parameters~\citep{xue2013restructuring}. However, there is no guarantee that the training procedure converges to a low rank weight matrix. Other works iterate between pruning and retraining of a neural network as a post processing step \citep{liu2015sparse,sun2016sparsifying,aghasi2017net,louizos2017bayesian}. While compressing a neural network by removing parameters  leads to a sparser neural network, this does not necessarily lead to a sparser representation. 
Indeed, a weight vector can be highly sparse but spread among the different neurons. This reduces the effective size of a neural network, from a compression point of view, but it would not be beneficial for later tasks as most of the neurons are already occupied by the current set of tasks. In our experiments, we show the difference between using a sparse penalty on the representation versus applying it to  the weights.
\section{Selfless Sequential Learning}
\label{sec:method}
\label{sec:method:ssl}
One of the main challenges in single model sequential learning is to have capacity to learn new tasks and at the same time avoid 
catastrophic forgetting of previous tasks as a result of learning new tasks. In order to prevent catastrophic forgetting, importance weight based methods such as 
EWC \citep{kirkpatrick2016overcoming} or
MAS \citep{aljundi2017memory} 
introduce
an importance weight $\Omega_{k}$ for each parameter $\theta_{k}$ in the network.
While these methods differ in how to estimate the important parameters, 
all of them 
penalize changes to important parameters  when learning a new task $T_n$  using $L_2$ penalty: 
\begin{equation} 
T_n:\quad \min_{\theta} \frac{1}{M} \sum_{m=1}^{M}  \mathcal{L}(y_m,f(x_m,\theta^n))+ \lambda_{\Omega}\sum_{k} \Omega_{k}(\theta^n_{k}-\theta^{n-1}_{k})^2
\end{equation}
where $\theta^{n-1}=\{\theta^{n-1}_{k}\}$ are the optimal parameters learned so far, i.e. before the current task.  $\{x_m\}$ is the set of $M$ training inputs,  with $\{f(x_m,\theta^n)\}$ and $\{y_m\}$ the corresponding predicted and desired outputs, respectively. 
$\lambda_{\Omega}$ is a trade-off parameter between the new  task objective {$\mathcal{L}$} and the changes on the important parameters, i.e. the amount of forgetting.

In this work we introduce an additional regularizer $R_{\textrm{SSL}}$ which encourages sparsity in the activations $H_l=\{h^m_{i}\}$ for each layer $l$.
\begin{equation}
\label{eq:appraoch:loss:final}
T_{n}: \quad \min_{\theta} \frac{1}{M} \sum_{m=1}^{M} \mathcal{L}(y_m,f(x_m,\theta^n))+ \lambda_{\Omega}\sum_{k} \Omega_{k}(\theta^n_{k}-\theta^{n-1}_{k})^2+\lambda_{\textrm{SSL}} \sum_{l} R_{\textrm{SSL}}(H_l)
\end{equation} 
 $\lambda_{\textrm{SSL}}$ and  $\lambda_{\Omega}$ are trade-off parameters that control the contribution of each term.
When training the first task ($n=1$),   $\Omega_{k}=0$. %

\subsection{\SNIFull}
\label{sec:method:sni}
Now we describe how we obtain a sparse and decorrelated representation. In the literature sparsity has been proposed by~\cite{pmlr-v15-glorot11a} to be combined with the rectifier activation function (ReLU) to control unbounded activations and to increase sparsity.
They minimize the  $L_1$ norm of the activations (since minimizing the  $L_0$ norm is an NP hard problem).  However,  $L_1$ norm imposes an equal penalty on all the active neurons leading to small activation magnitude across the network.\\
Learning a decorrelated representation has been explored before  with the goal of reducing overfitting. This is usually done by minimizing the Frobenius norm of the covariance matrix corrected by the diagonal, as in \cite{cogswell2015reducing} or \cite{xiong2016regularizing}. Such a penalty results in a decorrelated representation but with activations that are mostly close to a non zero mean value.
We merge the two objectives of sparse and decorrelated representation resulting in
the following objective:
\begin{equation}
\label{eq:appraoch:regularizer:simple}
R_{\SNIwoNI}(H_l)=\frac{1}{M}  \sum_{i,j}\sum_m  h_i^mh_j^m  ,  \quad i\neq j
\end{equation}
where we consider a hidden layer $l$ with activations $H_l=\{h^m_{i}\}$ for 
a set of inputs $X=\{x_m\}$
and $i,j \in {1,..,N}$ 
running over all $N$ neurons in the hidden layer.
 This formula differs from minimizing the Frobenius norm of the covariance matrix  in two simple yet important aspects:\\
 (1) In the case of a ReLU activation function, used in most modern architectures, a neuron is active  if its output is larger than zero, and zero otherwise. By assuming a close to zero mean of the 
 activations, $ \mu_i \simeq 0 \, \forall i \in {1,..,N}$, we minimize the 
 correlation between any two active neurons.\\
(2) By evaluating the derivative of the presented regularizer w.r.t. the activation, we get:
\begin{equation}
\frac {\partial R_{\SNIwoNI}(H_l)}{\partial h_i^m}=\frac{1}{M}   \sum_{j\neq i} h_j^m
\end{equation}
i.e., each active neuron receives a penalty from every other active neuron that corresponds to that other neuron's activation magnitude. In other  words, if a neuron fires, with a high activation value, for a given example, it will suppress firing of other neurons for that same example. Hence, this results in a decorrelated sparse representation.
\subsection{\SLNIFull}
\label{sec:method:SLNI}
The loss imposed by the %
$\SNIwoNI$
objective will only be zero when there is 
at most
one active neuron per example. This seems to be too harsh for complex 
tasks
that need a richer representation. Thus, we suggest to relax the objective by imposing a spatial weighting to the correlation penalty. 
In other words, an active neuron penalizes mostly its close neighbours and this effect vanishes for 
neurons further away.
Instead of uniformly penalizing all the correlated neurons, 
we weight the correlation penalty between  two neurons  with locations $i$ and $j$ %
using a Gaussian weighting. 
 This gives
\begin{equation}
\label{eq:appraoch:regularizer:first}
R_{\SLNIwoNI}(H_l)=
\frac{1}{M} \sum_{i,j} e^{-\frac{(i-j)^2}{2\sigma^{2}}}\sum_m  h_i^mh_j^m  ,  \quad i\neq j
\end{equation}
As such, each active neuron inhibits its neighbours,
introducing a locality in the network inspired by biological neurons. While the notion of neighbouring neurons is not well established in a fully connected network, our aim is to allow few  neurons to be active and not only one, thus those few activations don't have to be small to compensate for the penalty.
$\sigma^2$ is a hyper parameter representing the scale at which neurons can affect each other. Note that this is somewhat more flexible than decorrelating neurons in fixed groups as used in~\cite{xiong2016regularizing}.
Our regularizer inhibits locally the  active neurons leading to a sparse coding through local neural inhibition. 
\subsection{Neuron importance for discounting inhibition}
Our regularizer is to be applied for each task in the learning sequence. In the case of  tasks with completely different input patterns, the active neurons of the previous tasks will not be activated given the new tasks input patterns. However, when the new tasks are of similar
or shared patterns, neurons used for previous tasks will be active. In that case, our penalty would discourage other neurons from being active and encourage the new task to adapt the already active neurons instead. This would interfere with the previous tasks and could increase forgetting which is exactly what we want to overcome. To avoid such interference, we  add a weight factor taking into account the importance of the neurons with respect to the previous tasks. To estimate the importance of the neurons, we use as a measure the sensitivity of the loss at the end of the training to their changes. This is approximated by the gradients of the loss w.r.t. the neurons outputs (before the activation function) evaluated at each data point.
To get an importance value, we then accumulate the absolute value of the gradients over the given data points %
obtaining importance weight $\alpha_{i}$ for neuron $n_{i}$: 
\begin{equation}\label{eq:2}
\alpha_{i} =\frac{1}{M}\sum_{m=1}^M  \mid g_{i}(x_m) \mid \quad,\quad g_{i}(x_m)=\frac{\partial (\mathcal{L}(y_m,f(x_m,\theta^n)))}{\partial n^{m}_{i}}
\end{equation} 
 where $n^{m}_{i}$ is the output of  neuron $n_i$ for a given input example $x_m$, and $\theta^n$ are the parameters after learning task $n$. This is in line with the  estimation of the parameters importance in  \cite{kirkpatrick2016overcoming} 
 but considering the derivation variables to be the neurons outputs instead of the parameters.\\
Instead of relying on the gradient of the loss, we can also use the gradient of the learned function, i.e. the output layer, as done in \cite{aljundi2017memory} for estimating the parameters importance. During the early phases of this work, we experimented with both and observed a similar behaviour. For sake of consistency and computational efficiency 
we utilize the gradient of the function when using~\cite{aljundi2017memory} as LLL method and the gradient of the loss when experimenting with EWC~\citep{kirkpatrick2016overcoming}.
Then, we can weight our regularizer as follows:
\begin{equation} 
\label{eq:appraoch:regularizer:final}
R_{\SLNI}(H_l)=\frac{1}{M}   \sum_{i,j}e^{-(\alpha_{i}+\alpha_{j})} e^{-\frac{(i-j)^2}{2\sigma^{2}}} \sum_m h_i^mh_j^m ,  \quad i\neq j
\end{equation}
which can be read as: if  an important neuron for a previous task is active given  an input pattern from the current task, it will not suppress the other neurons from being active neither be affected by other active neurons.
For all other active neurons, local inhibition is deployed.
The final objective for training is given in Eq. \ref{eq:appraoch:loss:final}, setting $R_{SSL}:=R_{\SLNI}$ and $\lambda_{\textrm{SSL}}:=\lambda_{\SLNI}$. We refer to our full method as \SLNIDFull.

\section{Experiments}\label{sec:experiments}
In this section we study the 
 role of 
 standard
 regularization techniques with a focus on sparsity and decorrelation of the representation 
in a sequential learning scenario.  We first 
compare different activation functions and regularization techniques, including our proposed \SLNI,  on permuted MNIST (Sec. \ref{sec:experiments:analysis}). 
Then, we compare the top competing techniques and our proposed method in the case of sequentially learning  CIFAR-100 classes and 
Tiny Imagenet classes (Sec. \ref{sec:experiments:cifarimagenet}).  
Our \SLNI regularizer can be integrated in any importance weight-based lifelong learning approach such as \citep{kirkpatrick2016overcoming,zenke2017improved,aljundi2017memory}. Here we focus on Memory Aware Synapses~\citep{aljundi2017memory} ({\tt  MAS}), which is easy to integrate and experiment with and has shown superior performance~\citep{aljundi2017memory}. However, we also show results with Elastic weight consolidation~\citep{kirkpatrick2016overcoming}({\tt  EWC}) in Sec. \ref{sec:experiments:EWC}. Further, we ablate the components of our regularizer, {both in the standard setting (}Sec. \ref{sec:experiments:ablation})
{as in a setting without hard task boundaries (Sec.~\ref{sec:experiments:noboundaries})}.
Finally, we show how our regularizer improves the state-of-the-art performance on a sequence of object recognition tasks
(Sec. \ref{sec:experiments:object}).

\subsection{An in-depth comparison of regularizers and activation functions for Selfless Sequential Learning}
\label{sec:experiments:analysis}
\begin{figure*}[t]
\centering
\vspace*{-1.1cm} 
\includegraphics[width=0.92\textwidth]{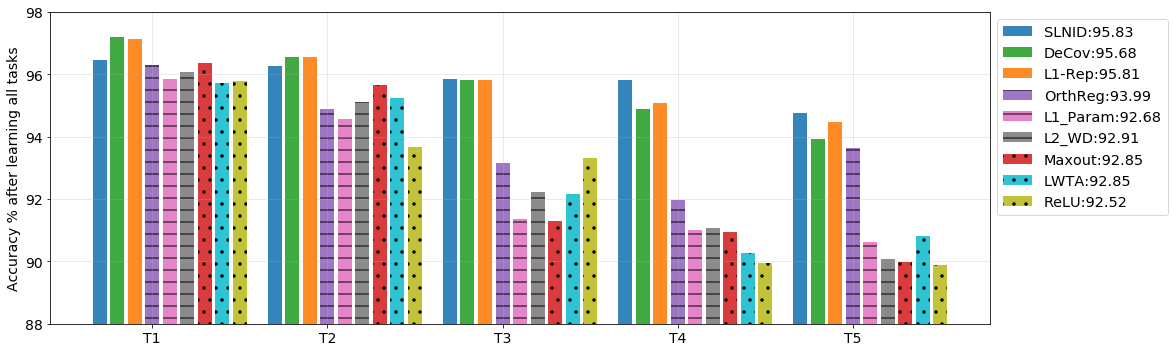}
\vspace*{-0.3cm} 
     \caption{\footnotesize Comparison of different regularization techniques on 5 permuted MNIST sequence. Representation based regularizers are solid bars, bars with lines represent parameters regularizers, dotted bars represent activation functions. Average test accuracy over all tasks is given in the legend. Representation based regularizers achieve  higher performance than other compared methods including parameters based regularizers. Our regularizer, \SLNI, performs the best on the last two tasks indicating that more capacity is left to learn these tasks.}%
    \label{fig:perm_comp}%
    \vspace*{-0.4cm} 
\end{figure*}
We study possible regularization techniques that could lead to less interference between the different tasks in a sequential learning scenario either by enforcing sparsity or decorrelation. Additionally, we examine the use of  activation functions that are inspired by lateral inhibition in biological neurons that could be advantageous in sequential learning.  MAS \cite{aljundi2017memory} is used in all cases as LLL method.\\
\textbf{Representation Based methods:} \\
-~{\tt L1-Rep}: To promote representational sparsity, an $L_1$ penalty on the activations is used.\\
-~{\tt Decov}~\citep{cogswell2015reducing}   aims at reducing overfitting by decorrelating  neuron activations. To do so, it minimizes the Frobenius norm of the  covariance matrix computed on the activations of the current  batch after subtracting the diagonal to avoid  penalizing independent  neuron activations. \\
\textbf{Activation functions:} \\
- {\tt Maxout} network~\citep{goodfellow2013maxout} utilizes the maxout activation function. For each group of neurons, based on a fixed window size, only the maximum activation is forwarded to the next layer. The activation function guarantees a minimum sparsity rate defined by the window size.  \\
- {\tt LWTA}~\citep{NIPS2013_5059}: similar idea to the Maxout network except that the non-maximum activations are set to zero while maintaining their connections. In contrast to Maxout, LWTA keeps the connections of the inactive neurons which can be occupied later once they are activated without changing the previously active neuron connections.\\
- {\tt ReLU}~\citep{pmlr-v15-glorot11a} The rectifier activation function (ReLU)  used as a baseline here and indicated in later experiments as {\tt No-Reg} as it represents the standard setting of sequential learning on networks with ReLU.  All the studied regularizers use ReLU as activation function.\\
\textbf{Parameters based regularizers:} \\
- {\tt OrthReg}~\citep{rodriguez2016regularizing}: Regularizing CNNs with locally constrained decorrelations. It aims at decorrelating the feature detectors  by minimizing the cosine of the angle between the weight vectors resulting eventually in orthogonal weight vectors.\\
- {\tt L2-WD}: Weight decay with $L_2$ norm ~\citep{krogh1992simple} controls the complexity of the learned function by minimizing the magnitude of the weights.\\
- {\tt L1-Param}:  $L_1$ penalty on the parameters  to encourage a solution with sparse parameters.

Dropout is not considered as its role contradicts our goal. While dropout can improve each task performance and reduce overfitting, it acts as a model averaging technique.   By randomly masking neurons, dropout forces the different neurons to work independently. As such it encourages a redundant representation. As shown by~\citep{goodfellow2013empirical} the best network size for classifying  MNIST digits when using dropout was  about $50\%$ more than without it. Dropout steers the learning of a task towards occupying a good portion of the network capacity, if not all of it, which contradicts the sequential learning needs.

\begin{wrapfigure}{r}{0.35\textwidth}
 \vspace*{-1cm} 
     \centering
     \vspace*{-0.7cm} 
      \subfloat{{\includegraphics[width=0.4\textwidth]{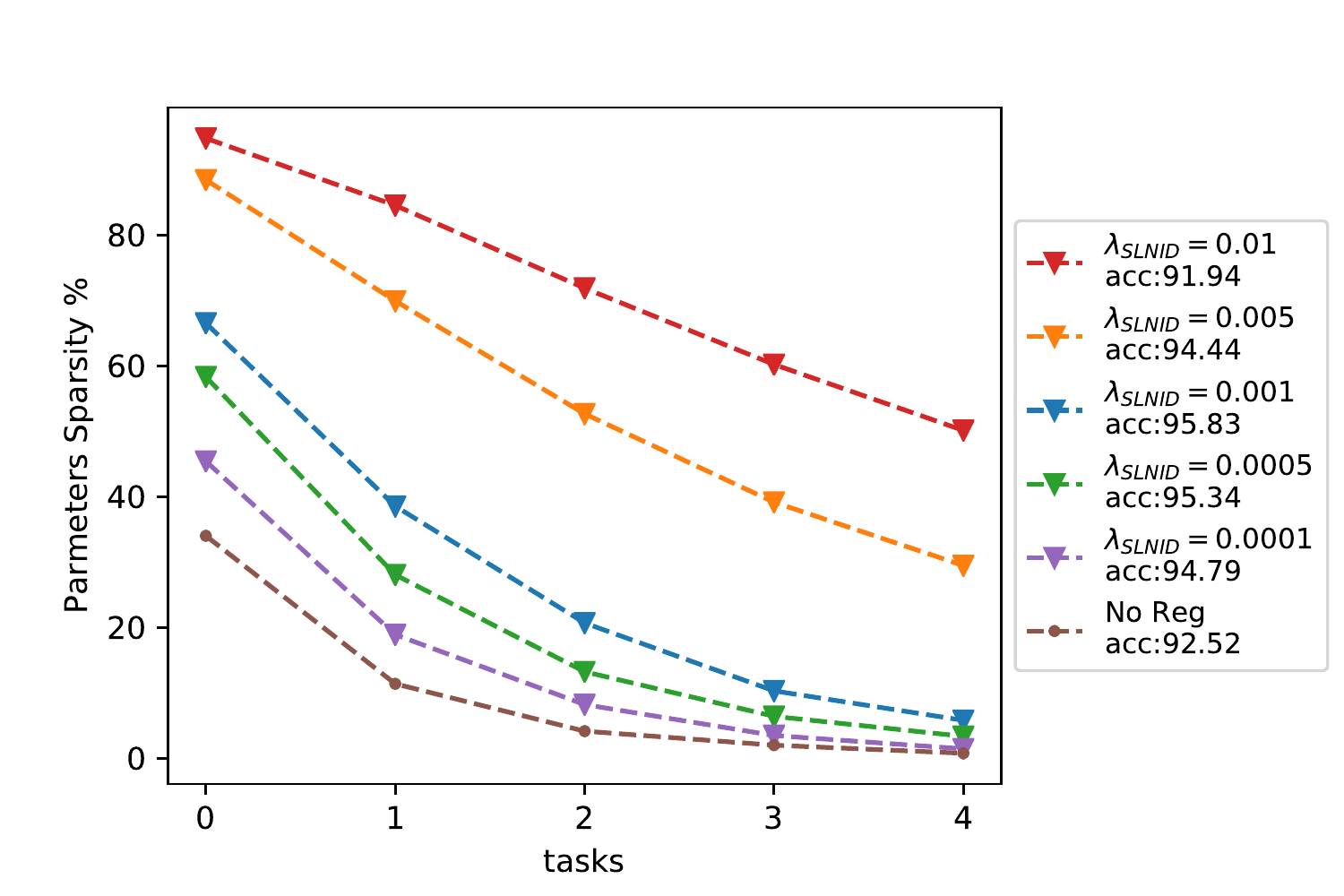} }}%
       \vfill
        \vspace*{-0.6cm}  
     \subfloat{{\includegraphics[width=0.35\textwidth]{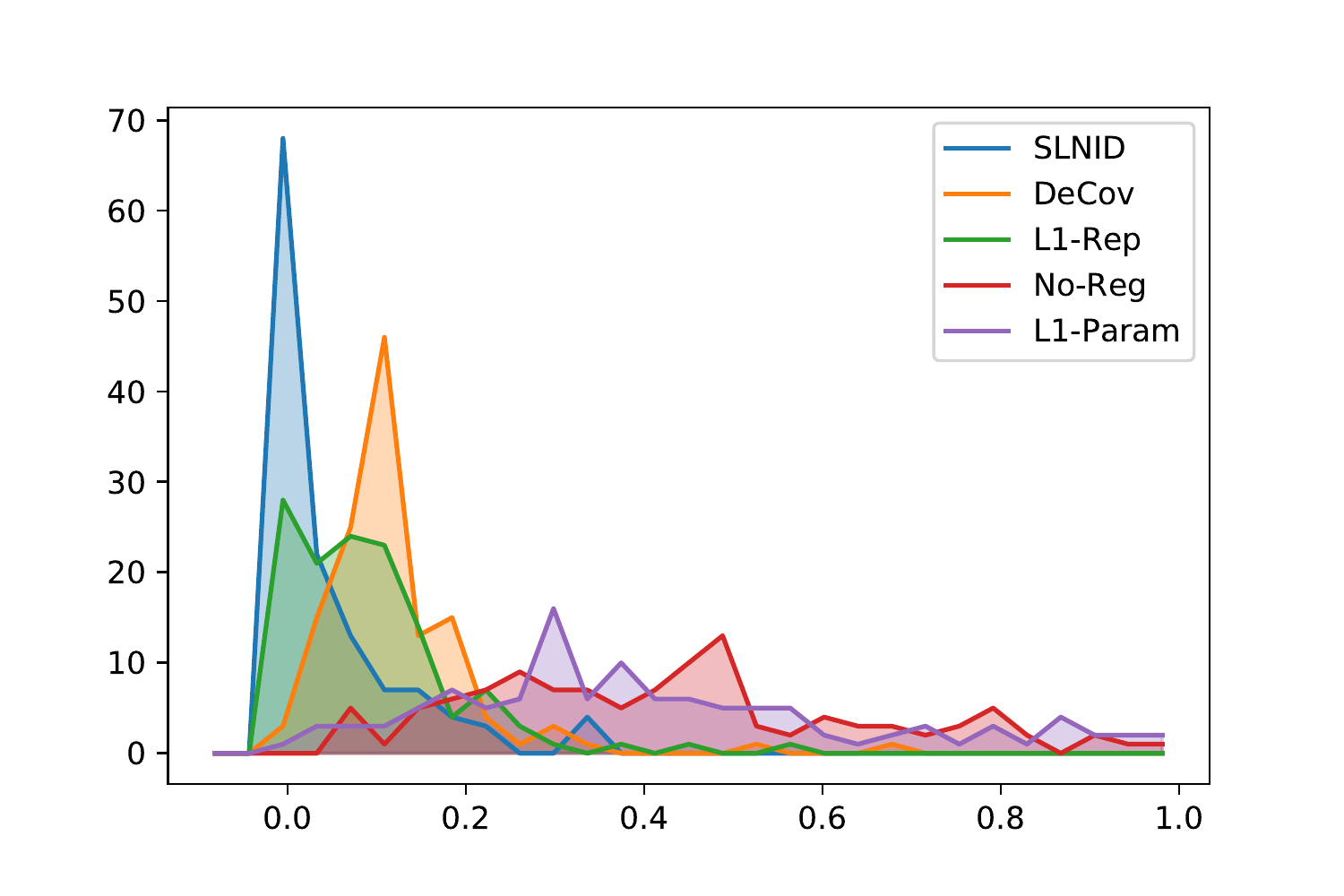} }}%
  \vspace*{-0.5cm} 
    \caption{\label{fig:hist_sparsity} \footnotesize  On the 5 permuted MNIST sequence, hidden layer=128, Top: percentage of unused parameters in the 1st layer using different $\lambda_{\SLNI}$; 
    Bottom:  histogram of neural activations on the first task. }
    \vspace*{-0.3cm}
\end{wrapfigure}

\myparagraph{Experimental setup.}
We use the MNIST dataset \citep{lecun1998gradient} as a first task in a sequence of 5  tasks, where we randomly permute  all the input pixels differently for  tasks 2 to 5. The goal is to classify MNIST digits from all the different permutations. 
The complete random permutation of the pixels in each task requires the neural network to instantiate a new neural representation for each pattern.
A similar setup has been used by~\cite{kirkpatrick2016overcoming,zenke2017improved,goodfellow2013empirical} with different percentage of permutations or different number of tasks.  \\
As a base network, we employ a multi layer perceptron with two hidden layers and a Softmax loss.  We experiment with  different number of neurons in the hidden layers $\{128,64\}$. 
For \SLNI we evaluate the effect of  $\lambda_{\SLNI}$ on the performance and the obtained sparsity
in Figure~\ref{fig:hist_sparsity}.
In general, the best $\lambda_{\SLNI}$ is the minimum value that maintains similar or better accuracy on the first task compared to the unregularized case, and we suggest to use this as a rule-of-thumb to set  $\lambda_{\SLNI}$.  
For $\lambda_{\Omega}$, we have used a high $\lambda_{\Omega}$ value that ensures the least forgetting which allows us to test the effect on the later tasks performance. Note that better average accuracies can be obtained with tuned $\lambda_{\Omega}$.
Please refer to Appendix \ref{appendix:experimental-setup} for hyperparameters and other details.

\myparagraph{Results:}
Figure~\ref{fig:perm_comp} presents the test accuracy on each task at the end of the sequence, achieved by the different regularizers and activation functions on the network with hidden layer of size $128$. Results on a network with hidden layer size $64$ are shown in the Appendix~\ref{appendix:additional}.
Clearly, in all the different tasks, the representational regularizers show a superior performance to the other studied techniques. For the regularizers applied to the parameters,  {\tt L2-WD} and {\tt L1-Param}
do not exhibit a clear trend and do not systematically show an improvement over the use of the different activation functions only. While {\tt OrthReg} shows a consistently good performance, it is lower than what can be achieved by the representational regularizers.
It is worth noting the {\tt L1-Rep} yields superior performance over {\tt L1-Param}. This observation is consistent across different sizes of the hidden layers (in Appendix~\ref{appendix:additional}) and shows the advantage of encouraging sparsity in the activations compared to that in the parameters.
Regarding the activation functions, {\tt Maxout} and {\tt LWTA} achieve a slightly higher performance than {\tt ReLU}. We did not observe a significant difference between the two activation functions. However, the improvement over {\tt ReLU} is only moderate
and does not justify the use of
a fixed window size and special architecture design.
Our proposed regularizer \SLNI achieves high if not the highest performance in all the tasks and succeeds in having a stable performance. This indicates the ability of \SLNI to direct the learning process towards using minimum amount of neurons and hence more flexibility for upcoming tasks.

\begin{figure}%
\vspace*{-1.2cm}
    \centering
   
    \subfloat{{\includegraphics[width=0.495\textwidth]{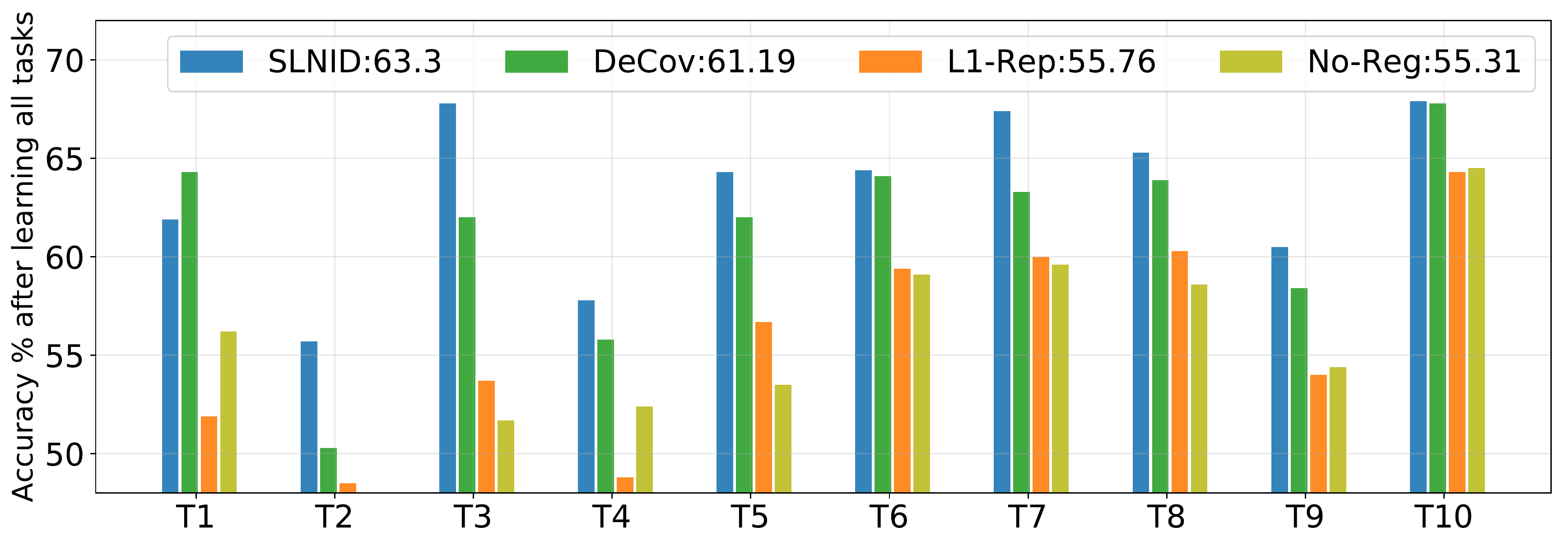} }}%
    \hfillx
    \subfloat{{\includegraphics[width=0.495\textwidth]{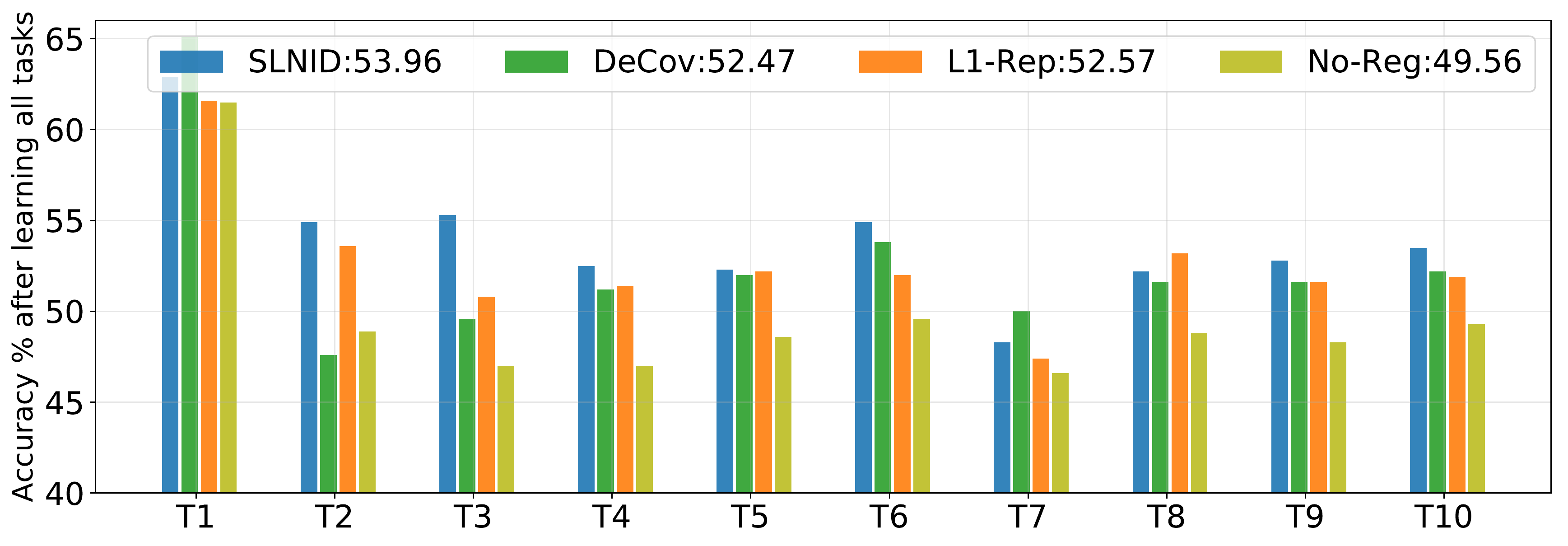} }}%
   \vspace*{-0.3cm} 
    \caption{\footnotesize  Comparison of different regularization techniques on a sequence of ten tasks from (a) Cifar split and (b)  Tiny ImageNet split. The legend shows average test accuracy over all tasks. Simple L1-norm regularizer (L1-Rep) doesn't help in such more complex tasks. Our regularizer \SLNI achieves an improvement of $2\%$ over Decov and $4-8 \% $ compared to No-Reg. }
    \label{fig:cifarimagenet}%
     \vspace*{-0.7cm}
\end{figure}

\myparagraph{Representation sparsity \& important parameter sparsity.} 

Here we want to examine the effect of our 
regularizer on the percentage of parameters that are utilized after each 
task and hence the
 capacity left for the later tasks. 
On the network with hidden layer size $128$, we compute the percentage of parameters with 
$\Omega_{k} < 10^{-2}$, with $\Omega_{k}$, see Appendix~\ref{appendix:experimental-setup}, 
the importance weight multiplier estimated and accumulated over tasks. 
Those parameters can be seen as unimportant and "free" for later tasks.  Figure~\ref{fig:hist_sparsity}(top)
shows the percentage of the unimportant (free) parameters in the first layer after each task for different $\lambda_{\SLNI}$ values along with the achieved average test accuracy at the end of the sequence.
It is clear that the larger $\lambda_{\SLNI}$, i.e., the more neural inhibition, the smaller the percentage of important parameters. Apart from the highest $\lambda_{\SLNI}$ where  tasks couldn't reach their top performance due to too strong inhibition,  improvement over the {\tt No-Reg} is always observed.
The optimal value for lambda seems to be the one that remains close to the optimal performance on the current task, while utilizing the minimum capacity feasible.
 Next, we compute the average activation per neuron, in the first layer, over all the examples and plot the corresponding histogram for \SLNI, {\tt DeCov}, {\tt L1-Rep}, {\tt L1-Param} and {\tt No-Reg} in Figure~\ref{fig:hist_sparsity}(bottom) at their setting that yielded the results shown in Figure~\ref{fig:perm_comp}. \SLNI has a peak at zero indicating  representation sparsity while the  other methods values are  spread along the line. 
This   seems to hint at the effectiveness of our  approach~\SLNI in learning a sparse yet powerful representation and in turn in a minimal interference 
 between tasks.

\adjustbox{valign=t*}{
  \begin{minipage}{\textwidth}
   \vspace*{0.5cm}
  \begin{minipage}[b]{0.5\textwidth}
  \tabcolsep=0.11cm   
  \footnotesize
  
\begin{tabular}{|l|l|l|l|l|}
\hline
 & \multicolumn{2}{|c|}{Permuted  mnist} & \multicolumn{2}{|c|}{Cifar}  \\ \hline
 h-layer dim. &   128 &   64 &   256 &  128  \\ \hline
 {\tt No-Reg} &92.67&90.72&55.06&55.3 \\ \hline
 $\SNIwoNI$ & 95.79 & {\bf 94.89} &  55.30 & 55.75 \\ 
 \SNI&  95.90&  93.82& 61.00& 60.90 \\ 
 $\SLNIwoNI$ & {\bf 95.95}  & 94.87  & 56.06  & 55.79 \\ 
 \SLNI&   95.83   & 93.89  &{\bf 63.30} & {\bf 61.16}\\ \hline
 Multi-Task Joint Training*&  97.30  & 96.80  &70.99  & 71.95\\ \hline
\end{tabular}
\captionof{table}{\label{tbl:ablation} \footnotesize  \SLNI ablation. Average test accuracy per task after training the last task in \%. * denotes that Multi-Task Joint Training violates the LLL scenario as it has access to all tasks at once and thus can be seen as an upper bound.  }
    \end{minipage}
    \hfillx
    \begin{minipage}[b]{0.4\textwidth}
      \footnotesize
      \vspace*{-1cm}
 \begin{tabular}{|l|l|}
\hline
Method&Avg-acc\\
\hline
Finetune&32.67\\
LWF~\citep{li2016learning}&49.49\\
 EBLL~\citep{triki2017encoder}& 50.29\\
IMM~\citep{lee2017overcoming}&43.4\\
Path Integral~\citep{zenke2017improved}&50.49\\
EWC~\citep{kirkpatrick2016overcoming}&50.00\\
MAS~\citep{aljundi2017memory}& 52.69\\
\SLNI -fc Pretrained (ours) &53.77\\
\SLNI -fc randomly initialized (ours) & {\bf 54.50}\\
\hline
\end{tabular}
    \captionof{table}{ \label{tbl:8tasks_obj} 8 tasks object recognition sequence. Average test accuracy per task after training the last task in \%.}
  \end{minipage}
 \end{minipage}}
\subsection{10 Task sequences on CIFAR-100 and Tiny Imagenet}
\label{sec:experiments:cifarimagenet}
While the previous section focused on learning a sequence of tasks with completely different input patterns and same objective, we now study the case of learning different categories of one dataset. 
For this we split the  CIFAR-100 and the Tiny ImageNet~\citep{yao2015tiny} dataset into ten tasks, 
respectively. We have 10 and 20 categories per task for CIFAR-100 and Tiny ImagNet, respectively. 
Further details about the experimental setup can be found in appendix  \ref{appendix:experimental-setup}.

We compare  the top competing methods from the previous experiments, {\tt L1-Rep}, {\tt DeCov}  and our \SLNI, and  {\tt No-Reg} as a baseline, {\tt ReLU} in previous experiment. Similarly, MAS \cite{aljundi2017memory} is used in all cases as LLL method.
Figures~\ref{fig:cifarimagenet}(a) and \ref{fig:cifarimagenet}(b) show the  performance on each of the ten tasks at the end of the sequence.
For both datasets, we observe that our \SLNI  performs overall best. {\tt L1-Rep} and {\tt DeCov} continue to improve over the non regularized case {\tt No-Reg}. These results confirm our proposal on the importance of sparsity and decorrelation in sequential learning.

\subsection{\SLNI with EWC \citep{kirkpatrick2016overcoming} } 
\label{sec:experiments:EWC}
\marcus{I think it would be good to have a table for this section, otherwise it will get lost.}
We have shown that our proposed regularizer~\SLNI exhibits stable and superior performance on the different tested networks when using {\tt MAS} as importance weight preservation method. 
To prove the effectiveness of our regularizer regardless of the used importance weight based method, we have tested~\SLNI on the 5 tasks permuted MNIST sequence in combination with Elastic Weight Consolidation ({\tt EWC},\cite{kirkpatrick2016overcoming}) and obtained a boost in the average performance at the end of the learned sequence equal to $3.1\%$ on the network with hidden layer size 128 and a boost of  $2.8\%$ with hidden layer size 64. Detailed accuracies are shown in Appendix~\ref{appendix:additional}.
It is worth noting that with both~{\tt MAS}  and~{\tt EWC} our~\SLNI was able obtain better accuracy using a network with a $64$-dimensional hidden size than when training without regularization {\tt No-Reg} on a network of double that size ($128$), indicating that~\SLNI allows to use neurons much more efficiently. 

\subsection{Ablation Study}
\label{sec:experiments:ablation}
Our method can be seen as composed of three components: the neural inhibition, the locality relaxation and the neuron importance integration.  To study how these components perform individually, Table \ref{tbl:ablation} reports the average accuracy at the end of the Cifar 100 and permuted MNIST sequences for each variant, namely, \SNI without neuron importance ($\SNIwoNI$), \SNI, \SLNI without neuron importance ($\SLNIwoNI$) in addition to our full \SLNI regularizer.  As we explained in Section~\ref{sec:method}, when tasks have completely different input patterns, the neurons that were activated on the previous task examples will not fire for new task samples and exclusion of important neurons is not mandatory. However, when sharing is present between the different tasks, a term to prevent \SLNI from causing any interference is required. This is manifested in the reported results: for permuted MNIST, all the variants work nicely alone, as a result of the simplicity and the disjoint nature of this sequence. However, in the Cifar 100 sequence, the integration of the neuron importance in the \SNI and \SLNI regularizers exclude important neurons from the inhibition, resulting in a clearly better performance. The locality in \SLNI  improves the performance in the Cifar sequence, which suggests that a richer representation is needed and  multiple active neurons should be tolerated.
\subsection{Sequential Learning without hard task boundaries}
\label{sec:experiments:noboundaries}
In the previous experiments, we  considered the standard task based scenario as in~\citep{li2016learning,zenke2017improved,aljundi2017memory,dataovercoming}, where at each time step we receive a task along with its training data and a new classification layer is initiated for the new task, if needed. Here, we are  interested in a more realistic scenario where the data distribution shifts gradually without  hard task boundaries.

\begin{wraptable}{r}{5cm}
\begin{scriptsize}
\begin{tabular}{|l|l|}
\hline
 Method& Avg.acc -tasks models\\
\hline
{\tt No-Reg} \texttt{w/o MAS}&69.20\%\\
\SLNIwoNI \texttt{w/o MAS} &72.14\% \\
\SLNI \texttt{w/o MAS} &73.03\% \\
No-Reg &66.88\%\\
\SLNIwoNI &71.32\% \\
\SLNI  &72.33\% \\
\hline
\hline
 Method& Avg.acc-last model \\
 \hline
{\tt No-Reg} \texttt{w/o MAS}&65.15\%\\
\SLNIwoNI \texttt{w/o MAS} &63.54\% \\
\SLNI \texttt{w/o MAS} &70.75\% \\
 \hline
{\tt No-Reg} &66.33\%\\
\SLNIwoNI &64.50\% \\
\SLNI  &70.94\% \\
\hline
\end{tabular}
\caption{\label{tab:no_boundaries} \footnotesize No tasks boundaries test case on Cifar 100. Top block, avg. acc  on each group of classes using each group model. Bottom block, avg. acc. on each group at the end of the training.}
\end{scriptsize}
 \vspace*{-0.3cm}
\end{wraptable}
To test this setting, we use the Cifar 100 dataset. Instead of considering a set of 10 disjoint tasks each composed of 10 classes, as in the previous experiment (Sec.~\ref{sec:experiments:cifarimagenet}), we now start by sampling with high probability $(2/3)$ from the first 10 classes and with low probability $(1/3)$ from the rest of the classes.  We train the  network (same architecture as
in Sec.~\ref{sec:experiments:cifarimagenet}) for a few epochs and then change the sampling probabilities to be high $(2/3)$ for classes $11-20$  and low $(1/3)$ for the remaining classes. This process is repeated until sampling with high probability  from the last 10 classes and low from the rest.  We use one shared classification layer throughout and estimate the importance weights and the neurons importance after each training step (before changing the sampling probabilities). 
We consider 6 variants: our \SLNI, the ablations \SLNIwoNI and without regularizer {\tt No-Reg}, as in Section~\ref{sec:experiments:ablation}, as well each of these three trained without the MAS importance weight regularizer of~\cite{aljundi2017memory}, denoted as \texttt{w/o MAS}.
Table~\ref{tab:no_boundaries} presents the accuracy {averaged over} the ten groups of ten classes,  using each group model (i.e.~the model trained when this group was sampled with  high probability) in the top block and the average accuracy on each of the ten groups at the end of the training (middle and bottom block).
We can deduce the following:  1) \SLNI improves the performance considerably (by more than 4\%) even without importance weight regularizer. 
2) In this scenario without hard task boundaries there is less forgetting than in the scenario with hard task boundaries
studied in Section~\ref{sec:experiments:cifarimagenet} for Cifar (difference between rows in top block to corresponding rows in middle block). {As a result,} the improvement obtained by deploying the importance weight regularizer is moderate: 
at $70.75\%$, \SLNI \texttt{w/o MAS} is already better than {\tt No-Reg} reaching $66.33\%$. 3) While \SLNIwoNI without MAS improves the individual models performance ($72.14\%$ 
compared to $69.20\%$), it fails to improve the overall performance at the end of the sequence ($63.54\%$ compared to $65.15\%$), as important neurons are not excluded from the penalty and hence they are changed or inhibited leading to tasks interference and performance deterioration.  

\subsection{Comparison with the state of the art}
\label{sec:experiments:object}
To compare our proposed approach with the different state-of-the-art sequential learning methods, we use a sequence of 8 different object recognition tasks, introduced  in~\cite{aljundi2017memory}. The sequence starts from AlexNet~\citep{NIPS2012_4824} pretrained on ImageNet~\citep{ILSVRC15} as a base network, following the setting of \cite{aljundi2017memory}. More details are in Appendix~\ref{appendix:experimental-setup:objects}. We compare against the following:
\noindent
{\em Learning without Forgetting}~\citep{li2016learning} ({\tt LwF}), {\em Incremental Moment Matching}~\citep{lee2017overcoming} ({\tt IMM}), {\em Path Integral}~\citep{zenke2017improved} and sequential finetuning ({\tt FineTuning}),
in addition to the case of {\tt MAS} \citep{aljundi2017memory} alone,~i.e. our {\tt No-Reg}  before. Compared methods were run with the exact same setup as in \cite{aljundi2017memory}. For our regularizer, we disable  dropout, since dropout encourages redundant activations which contradicts  our regularizer's role. Also, since the network is pretrained, the locality introduced in \SLNI may conflict with the already pretrained activations. For this reason, we also test \SLNI with randomly initialized fully connected layers. Our regularizer is applied with {\tt MAS} as a sequential learning method.
Table~\ref{tbl:8tasks_obj} reports the average test accuracy at the end of the sequence achieved by each method. \SLNI improves even when starting from a pretrained network and disabling dropout. Surprisingly, even with randomly initialized fully connected layers, \SLNI improves $1.8\%$ over the state of the art using a fully pretrained network.  
\section{Conclusion}\label{sec:conclusion}
In this paper we study the problem of sequential learning using a network with fixed capacity 
-- a prerequisite for a scalable and computationally efficient solution.
A key insight of our approach is that in the context of sequential learning (as opposed to other contexts where sparsity is imposed, such as network compression or avoiding overfitting), sparsity should be imposed at the level of the representation rather than at the level of the network parameters. 
Inspired by lateral  inhibition in the mammalian brain, we impose sparsity by means of a new regularizer that decorrelates nearby active neurons. 
We integrate this in a model which \emph{learns selflessly} a new task by leaving capacity for future tasks and at the same time \emph{avoids forgetting} previous tasks by taking into account neurons importance.
\newline 
\small {\textbf{Acknowledgment:}
 The first author's PhD is funded by an FWO scholarship.
\bibliography{iclr2019_conference}
\bibliographystyle{iclr2019_conference}

\clearpage

\section*{Appendix}
\appendix
\section{Details on the experimental setup}
In all designed experiments, our regularizer is applied to the neurons of the fully connected layers. As a future work, we plan to integrate it in the convolutional layers.
\label{appendix:experimental-setup}
\subsection{Permuted Mnist}
The used network is composed of two fully connected layers. All tasks are trained for 10 epochs with a learning rate $10^{-2}$ using SGD optimizer. ReLU is used as an activation function unless mentioned otherwise.
Throughout the experiment, we used a scale $\sigma$ for the Gaussian function used for the local inhibition equal to $1/6$ of the hidden layer size. 
For all competing
regularizers, we tested different hyper parameters from $10^{-2}$ to $10^{-9}$ and report the best one. 
For $\lambda_{\Omega}$, we have used a high $\lambda_{\Omega}$ value that ensures the least forgetting.
This allows us to examine the degradation in the performance  on the later tasks compared to those learned previously as a result of lacking capacity. Note that better average accuracies can be obtained with tuned $\lambda_{\Omega}$.

In section~\ref{sec:experiments:analysis} we estimated the free capacity in the network with the percentage of $\Omega_{k} < 10^{-2}$, with $\Omega_{k}$, 
the importance weight multiplier estimated and accumulated over tasks. We consider $\Omega_{k} < 10^{-2}$ of negligible importanc as in a network trained without a sparsity regularizer, $\Omega_{ij} < 10^{-2}$ covers the first 10 percentiles. 
\subsection{CIFAR-100}
As a base network, we use 
a network similar to the one used by~\cite{zenke2017improved} but without dropout. We evaluate two variants with hidden size $N=\{256,128\}$.
Throughout the experiment, we again used a scale $\sigma$ for the Gaussian function  equal to $1/6$ of the hidden layer size. 
We train the different tasks  for 50 epochs with a learning rate of $10^{-2}$ using SGD optimizer.
\subsection{Tiny ImageNet}
 We split the Tiny ImageNet dataset~\citep{yao2015tiny} into ten tasks, each containing twenty categories to be learned at once. As a base network, we use a variant of VGG~\citep{simonyan2014very}.
 For architecture details, please refer to Table~\ref{tab:arch_imagenet} below.
 \begin{table}[h]
\centering
 \begin{tabular}{|l|l|}
\hline
Layer& \# filters/neurons\\
\hline
Convolution&64\\
Max Pooling&-\\
Convolution&128\\
Max Pooling&-\\
Convolution& 256\\
Max Pooling&-\\
Convolution& 256\\
Max Pooling&-\\
Convolution& 512\\
Convolution& 512\\
Fully connected & 500\\
Fully connected & 500\\
Fully connected & 20\\
\hline
\end{tabular}
\caption{Architecture of the network used in the Tiny Imagenet experiment.}
\label{tab:arch_imagenet}
 \end{table}

 Throughout the experiment, we again used a scale $\sigma$ for the Gaussian function equal to $1/6$ of the hidden layer size. 
 \subsection{8 task object recognition sequence}
\label{appendix:experimental-setup:objects}
 The 8 tasks sequence is composed of: 1. Oxford \textit{Flowers}~\citep{Nilsback08}, 2. MIT \textit{Scenes} \citep{quattoni2009recognizing}, 3. Caltech-UCSD \textit{Birds}~\citep{WelinderEtal2010}, 4. Stanford  {\tt Cars}~\citep{krause20133d}; 5. FGVC-{\tt Aircraft}~\citep{maji13fine-grained}; 6. VOC {\tt Actions}
~\citep{pascal-voc-2012}; 7. {\tt Letters}~\citep{deCampos09}; and 8. {\tt SVHN}~\citep{netzer2011reading} datasets. 
We have rerun the different methods and obtain the same reported results as in ~\cite{aljundi2017memory}.

\section{Extra Results}
\label{appendix:additional}
\subsection{Permuted Mnist Sequence}
In section \ref{sec:experiments:analysis}, we have studied the performance of different regularizers and activation functions on 5 permuted Mnist tasks in a network with a hidden layer of size $128$.
Figure~\ref{fig:perm_comp_64} shows the average accuracies achieved by each of the studied methods at the end of the learned sequence in a network with a hidden layer of size $64$. Similar conclusions can be drawn. {\tt Maxout} and {\tt LWTA} perform similarly
and improve slightly over {\tt ReLU}. Regularizers applied to the representation are more powerful for sequential learning than regularizers applied directly to the parameters. Specifically, {\tt L1-Rep} (orange) is consistently better than {L1-Param} (pink). Our \SLNI is able of maintaining a good performance on all the tasks, achieving among the top average test accuracies.  
Admittedly, the performances of \SLNI is very close to L1-Rep. The difference between these methods stands out more clearly for larger networks and more complex tasks.

\begin{figure*}[h]
\centering

\includegraphics[width=0.95\textwidth]{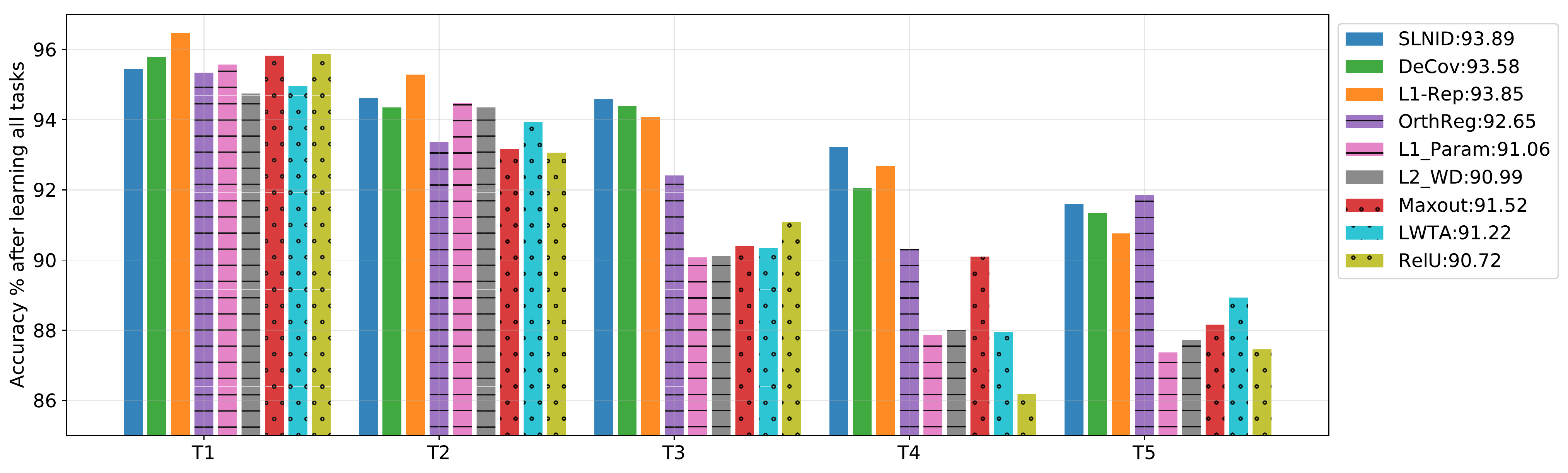} 
     \caption{\footnotesize Comparison of different regularization techniques on 5 permuted MNIST sequence of tasks, hidden size=64. Representation based regularizers are solid bars, bars with lines represent parameters regularizers, dotted bars represent activation functions. See Figure~\ref{fig:perm_comp} for size $128$.}%
    \label{fig:perm_comp_64}%

\end{figure*}
\subsection{SLNI with EWC}
To show that our approach is not limited to {\tt MAS}~\citep{aljundi2017memory}, we have also experimented with EWC~\citep{kirkpatrick2016overcoming} as another importance weight based method along with our regularize \SLNI on the permuted Mnist sequence.
Figure~\ref{fig:SLNI_EWC} shows the test accuracy of each task at the end of the 5 permuted Mnist sequence achieved by our \SLNI combined with {\tt EWC} and by {\tt No-Reg } (here indicating {\tt EWC} without regularization). It is clear that \SLNI succeeds to improve the performance on all the learned tasks which validates the utility of our approach with different sequential learning methods.
\begin{figure}[h]
\vspace*{-0.2cm}
    \centering
    \subfloat{{\includegraphics[width=0.495\textwidth]{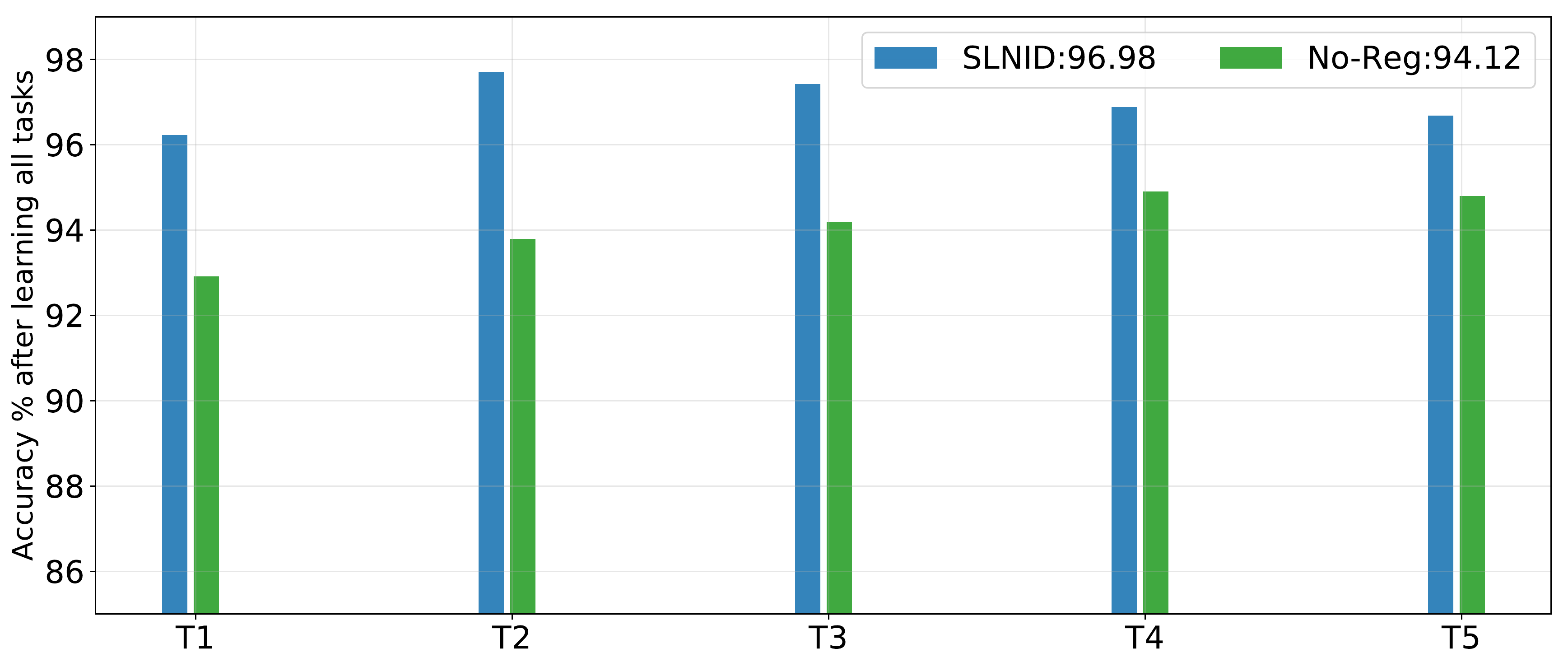} }}%
    \hfillx
    \subfloat{{\includegraphics[width=0.495\textwidth]{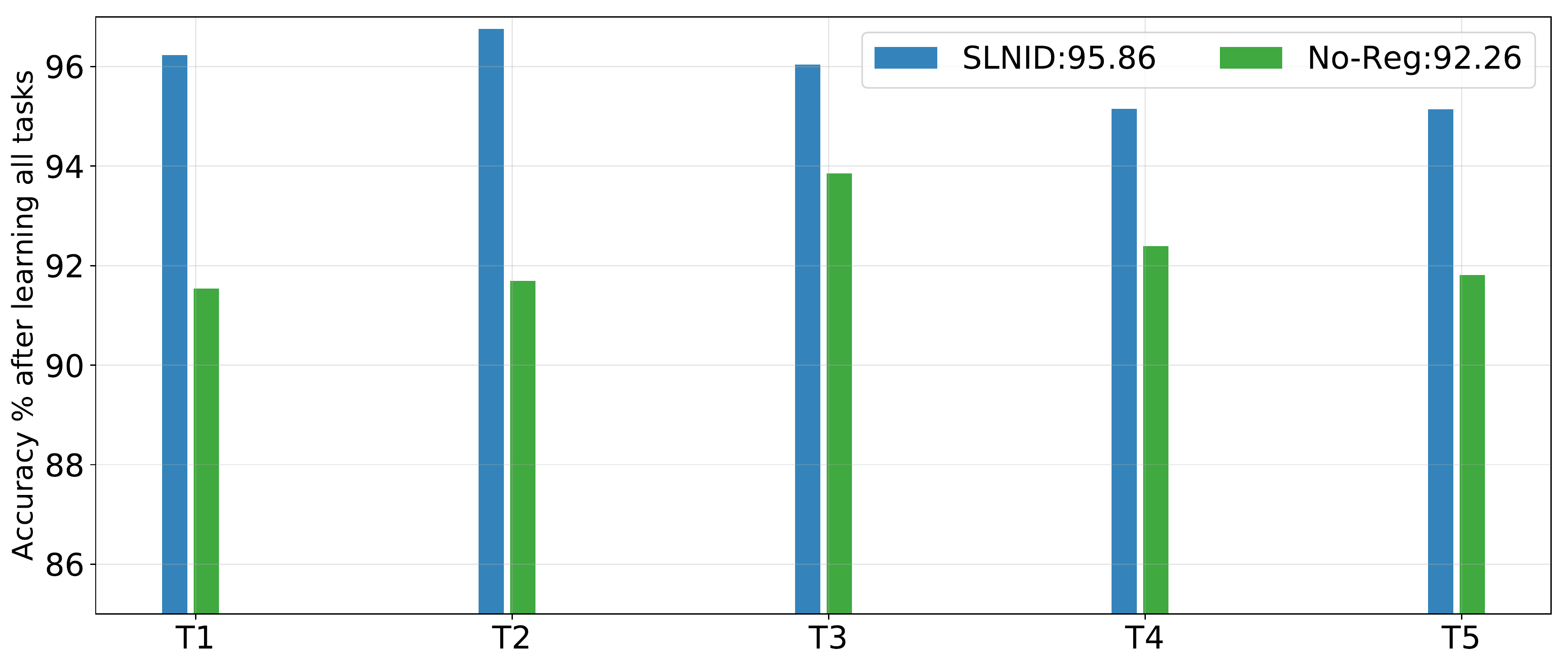} }}%
  
    \caption{\footnotesize (a) \SLNI with {\tt EWC} on 5 permuted Mnist sequence of tasks, hidden size=128, (b) hidden size=64. }
    \label{fig:SLNI_EWC}%
\end{figure}
\subsection{Cifar 100 sequence}
In section~\ref{sec:experiments:cifarimagenet} we have tested  \SLNI and other representation regularizers on the Cifar 100 sequence. In Figure~\ref{fig:cifarimagenet}(a) we compare their performance on a network with hidden layer size $256$. Figure~\ref{fig:cifar_128} 
repeats the same experiment for a network with hidden size $128$.  While {\tt DeCov} and \SLNI continue to improve over {\tt No-Reg}, {\tt L1-Rep} seems to suffer in this case. Our interpretation is that {\tt L1-Rep} here interferes with the previously learned tasks while penalizing activations and hence suffers from catastophic forgetting. In line with all the previous experiments \SLNI achieves the best accuracies and manages here to improve over {$6\% $} compared to {\tt No-Reg}.
\begin{figure*}[h!]
\centering
\includegraphics[width=0.80\textwidth]{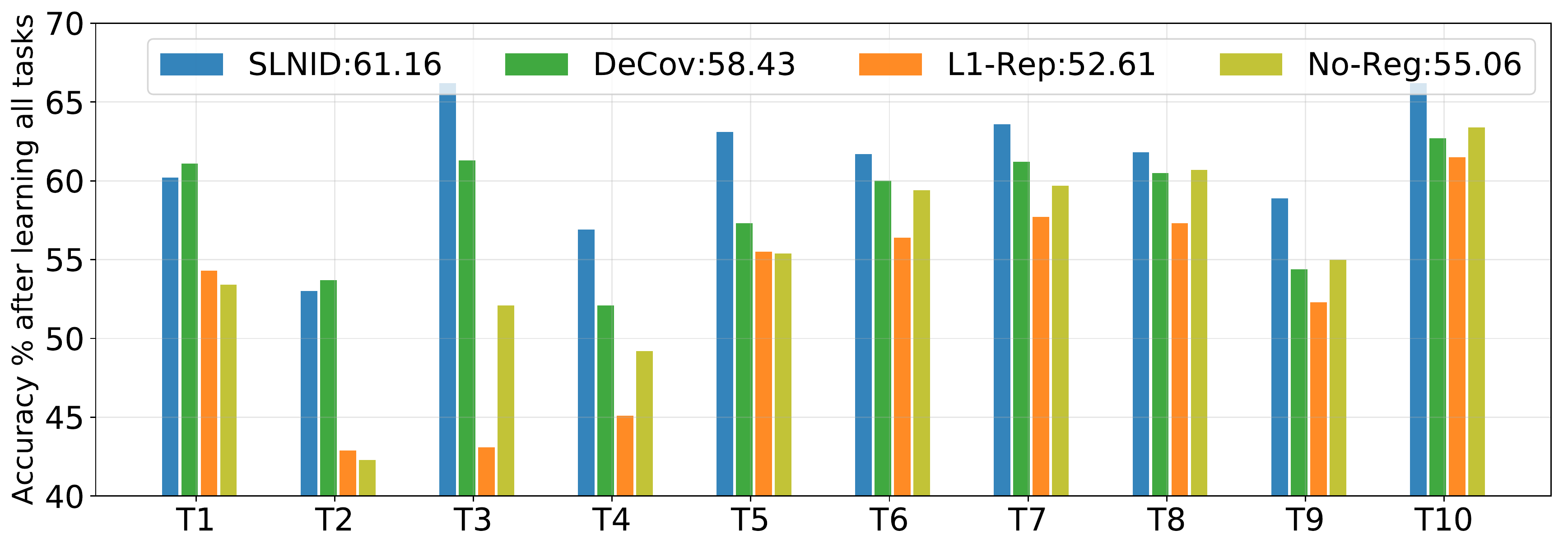} 
     \caption{\footnotesize  Comparison of different regularization techniques on a sequence of ten tasks from Cifar split. Hidden size=128. See Figure~\ref{fig:cifarimagenet}(a) for size 256.}%
    \label{fig:cifar_128}%
  
\end{figure*}

\subsection{Spatial Locality Test}
To avoid penalizing all the active neurons,  our \SLNI weights the correlation penalty between each two neurons based on their spatial distance using  a Gaussian function. We want to visualize the effect of this spatial locality on the neurons activity. To achieve this, we have used the first 3 tasks of the Permuted Mnist sequence as a test case and visualized the neurons importance after each task. This is done using the network of hidden layer size $64$. 
Figure~\ref{fig:scale1}, Figure~\ref{fig:scale2} and Figure~\ref{fig:scale3} show the neurons importance after each task. The left column is without locality, i.e. \SLNI, and the right column is \SLNI. Blue represents the first task, orange the second task and green the third task.
When using \SLNI, inhibition is applied in a local manner allowing more active neurons which could potentially improve the representation power. When learning the second task, new neurons become important regardless of their closeness to first task important neurons as those neurons are excluded from the inhibition. As such, new neurons are becoming active as new tasks are learned. For \SLNI all neural correlation is penalized in the first task. And for later tasks, very few neurons are able to become active and important for the new task due to the strong global inhibition, where previous neurons that are excluded from the inhibition are easier to be re-used.


  
  
  

\newpage
\begin{figure}[h]
\vspace*{-0.2cm}
    \centering
    \subfloat{{\includegraphics[width=0.495\textwidth]{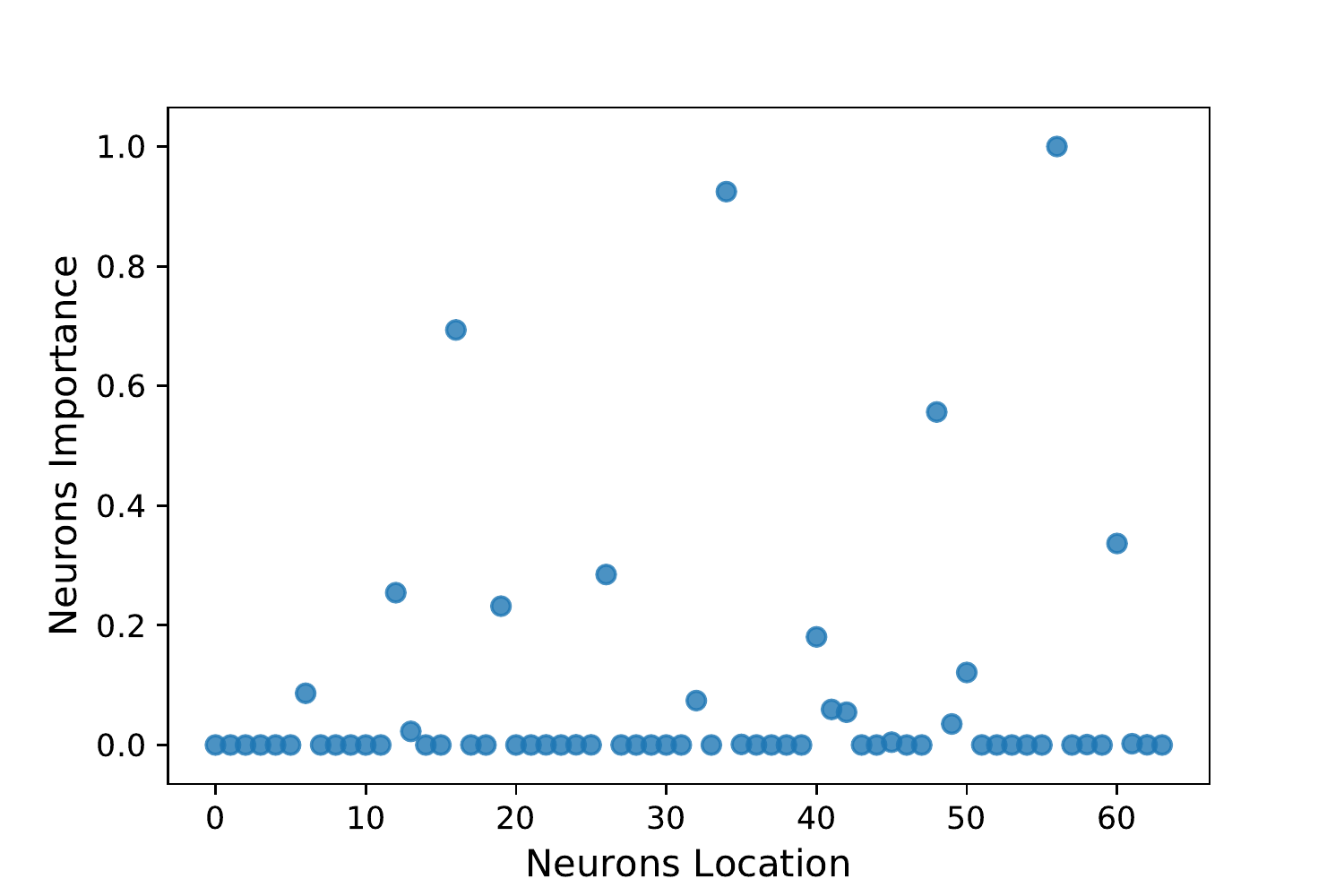} }}%
    \hfillx
    \subfloat{{\includegraphics[width=0.495\textwidth]{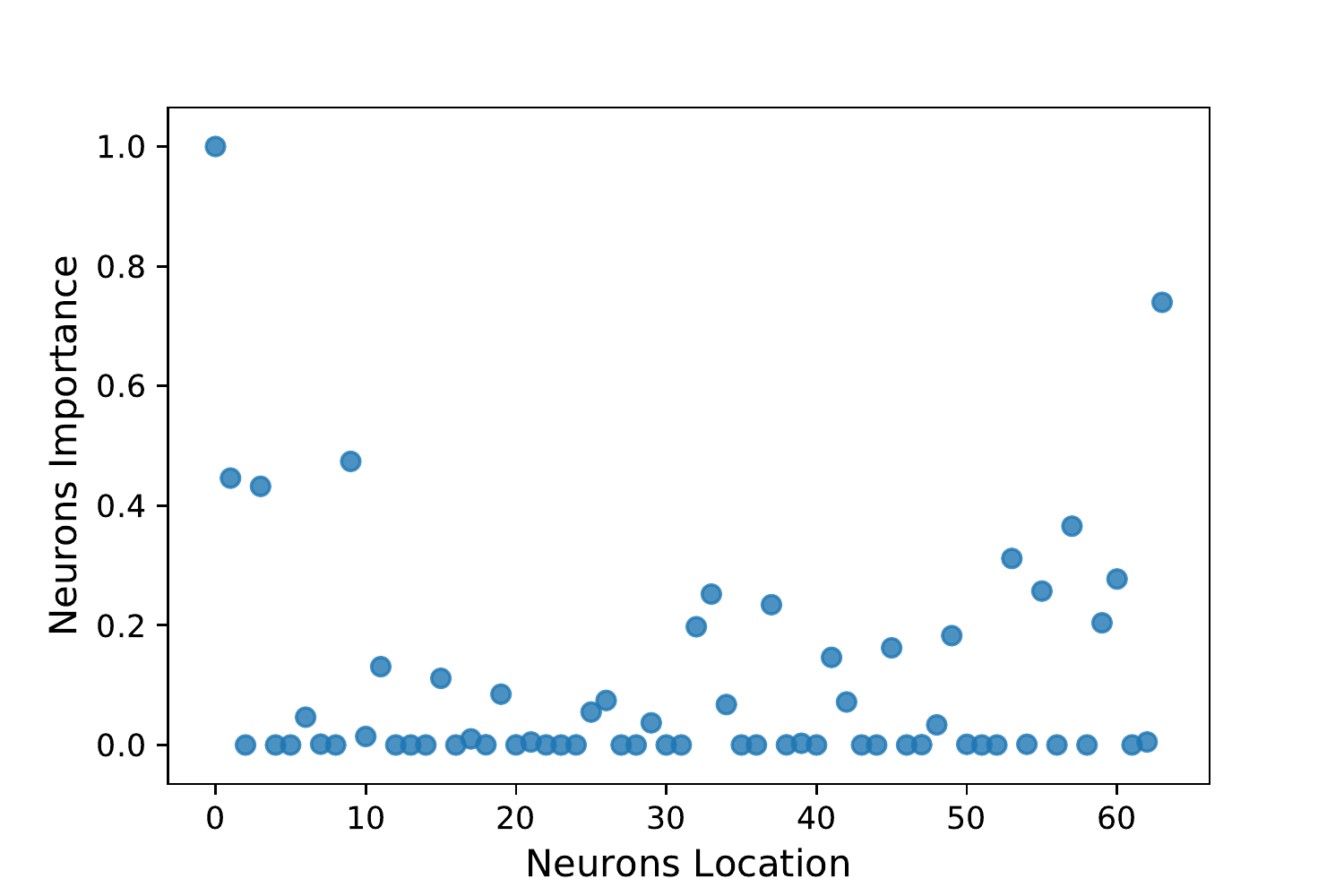} }}%
  
    \caption{\footnotesize First layer neuron importance after  learning the first task (blue). Left:  \SNI, Right:  \SLNI.  More active neurons are tolerated in  \SLNI.}
    \label{fig:scale1}%
\end{figure}

\begin{figure}[h]
\vspace*{-0.2cm}
    \centering
    \subfloat{{\includegraphics[width=0.495\textwidth]{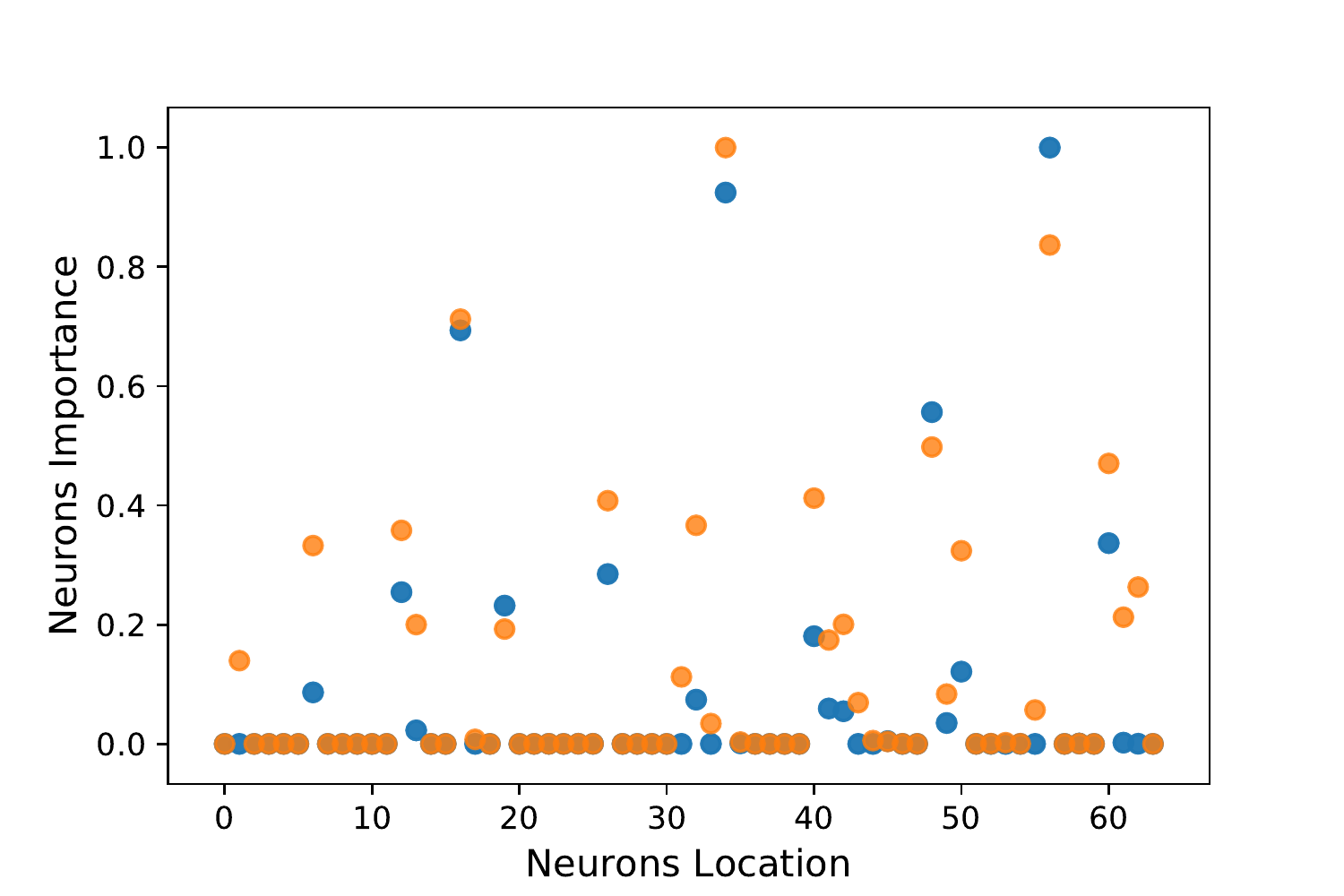} }}%
    \hfillx
    \subfloat{{\includegraphics[width=0.495\textwidth]{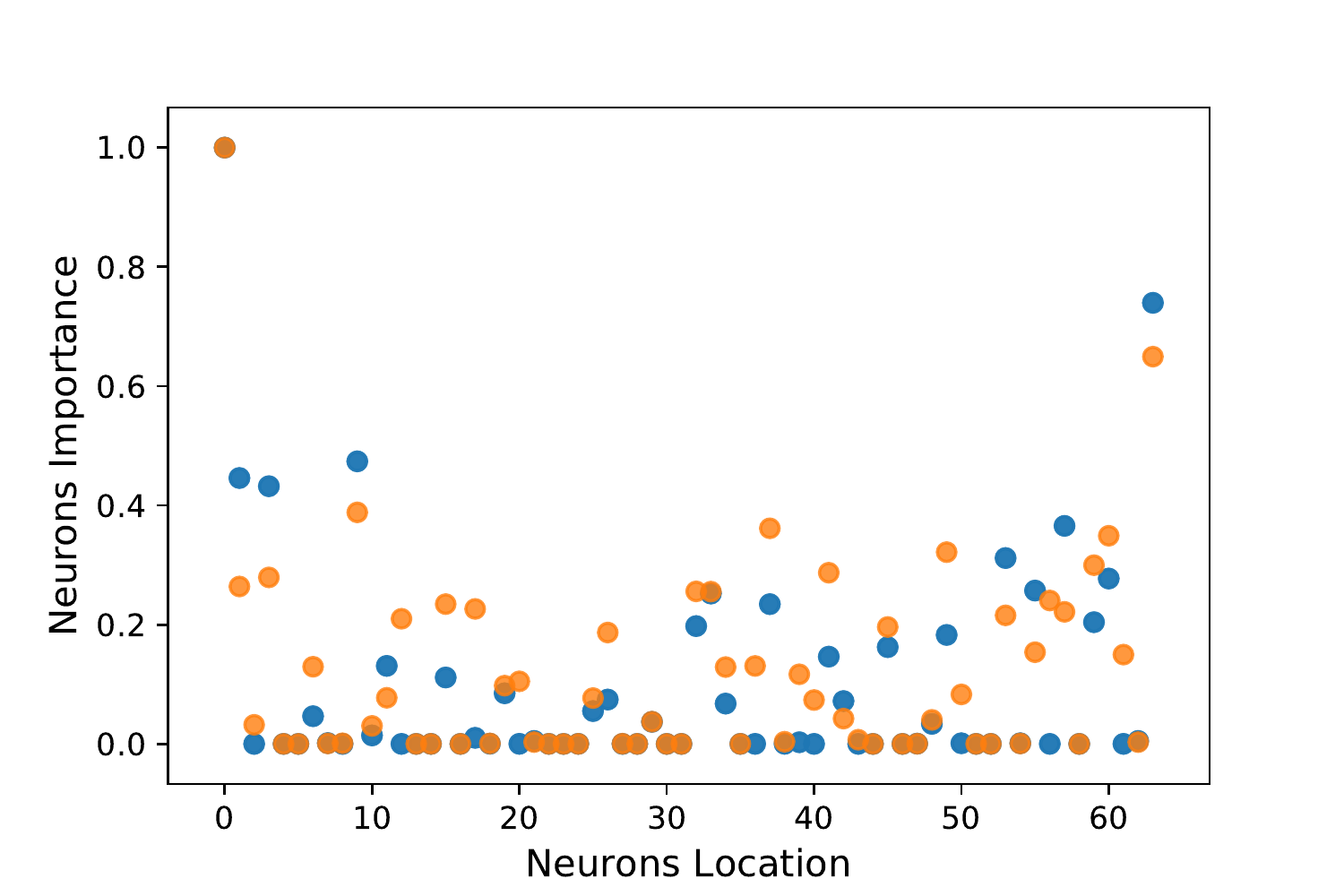} }}%
  
    \caption{\footnotesize First layer neuron importance  after learning the second task (orange), superimposed on Figure~\ref{fig:scale1}. Left: \SNI, Right: \SLNI. \SLNI allows new neurons, especially those that were close neighbours to previous important neurons, to become active and to be used for the new task. \SNI penalizes all unimportant neurons equally. As a result, previous neurons are adapted for the new tasks and less new neurons are getting activated. }
    \label{fig:scale2}%
\end{figure}
\begin{figure}[h]
\vspace*{-0.2cm}
    \centering
    \subfloat{{\includegraphics[width=0.495\textwidth]{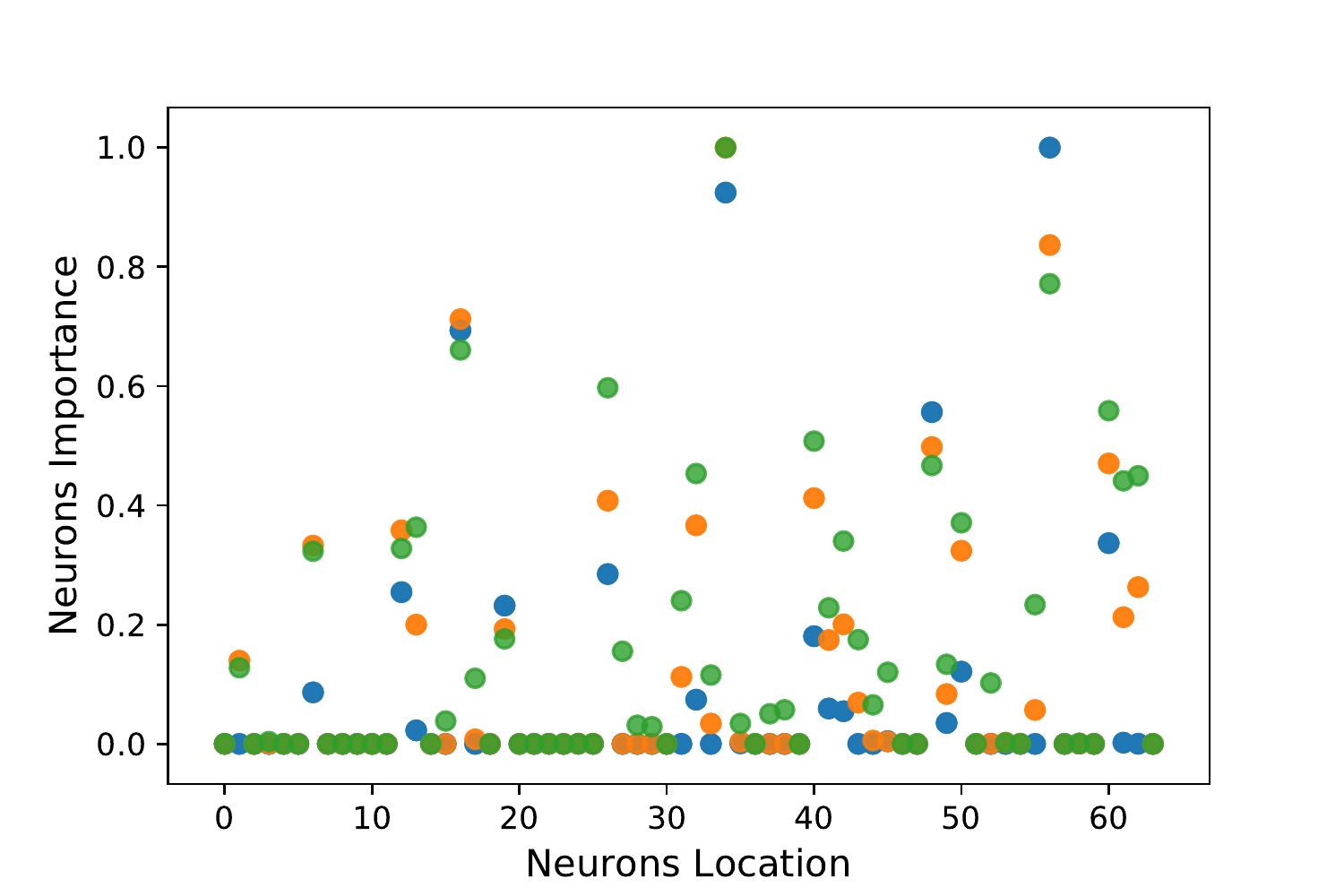} }}%
    \hfillx
    \subfloat{{\includegraphics[width=0.495\textwidth]{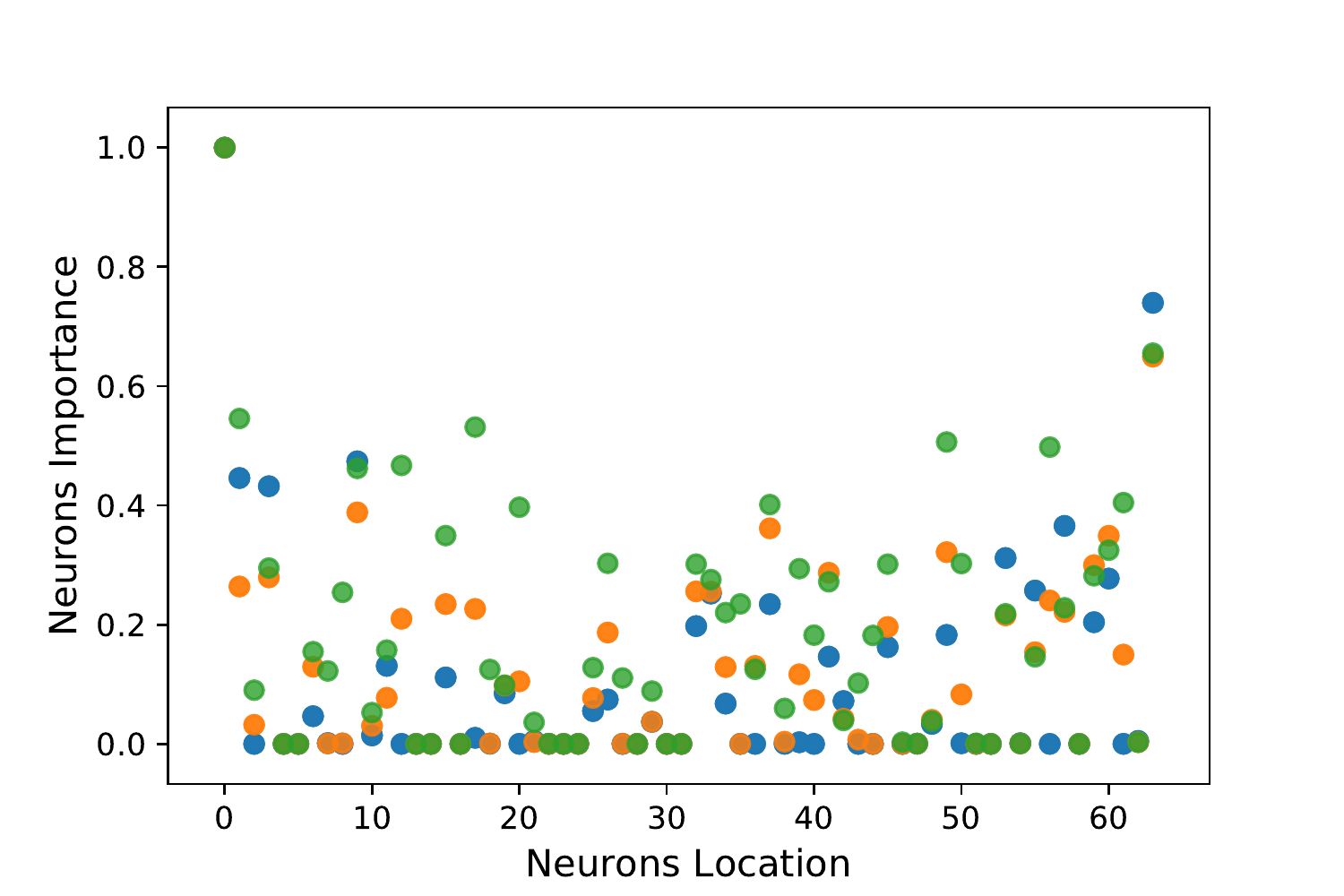} }}%
  
    \caption{\footnotesize First layer neuron importance after  learning the third task (green), superimposed on Figure~\ref{fig:scale2}. Left: \SNI, Right:  \SLNI.  \SLNI allows previous neurons to be re-used for the third task. It avoids changing the previous important neurons by adding new neurons. For \SNI, very few neurons are newly deployed. The new task is learned mostly by adapting previous important neurons, causing more interference. }
    \label{fig:scale3}%
\end{figure}
\newpage
\begin{figure}[h]
\vspace*{-0.2cm}
    \centering
    \subfloat{{\includegraphics[width=0.495\textwidth]{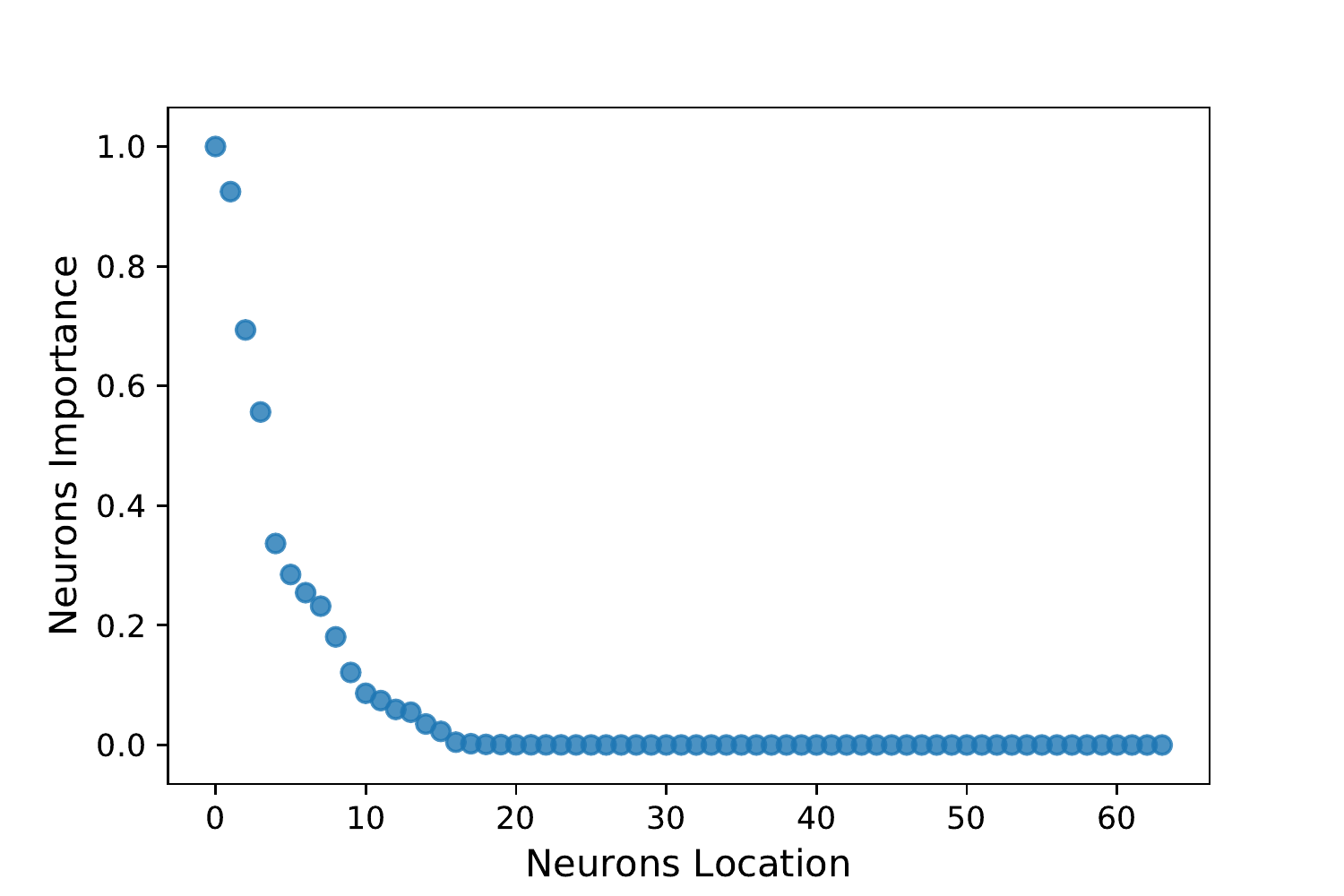} }}%
    \hfillx
    \subfloat{{\includegraphics[width=0.495\textwidth]{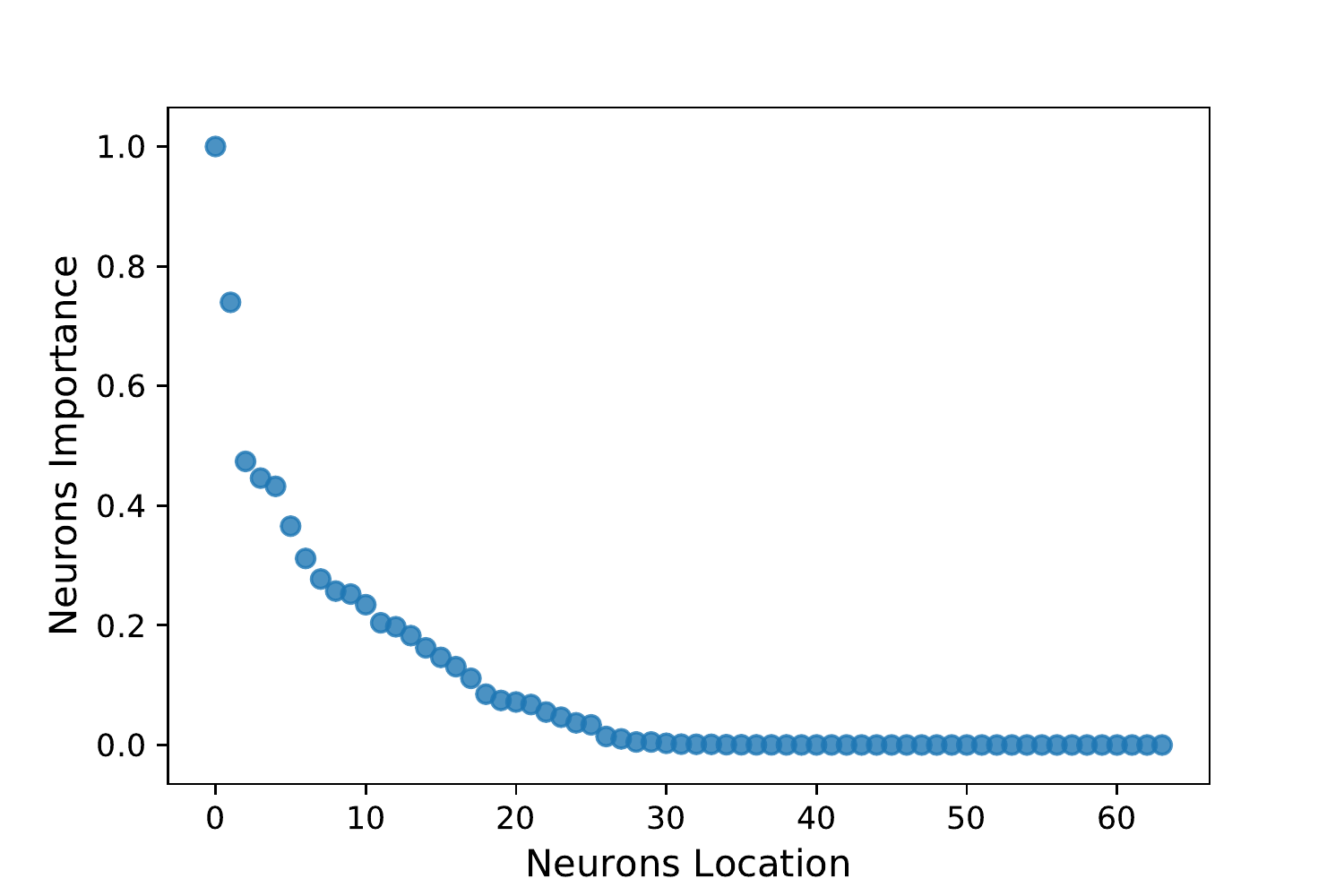} }}%
  
    \caption{\footnotesize First layer neuron importance after  learning the first task, sorted in descending order according to the first task neuron importance (blue). Left: \SNI, Right:  \SLNI.  More active neurons are tolerated in  \SLNI.}
    \label{fig:scale1b}%
\end{figure}
\begin{figure}[h]
\vspace*{-0.2cm}
    \centering
    \subfloat{{\includegraphics[width=0.495\textwidth]{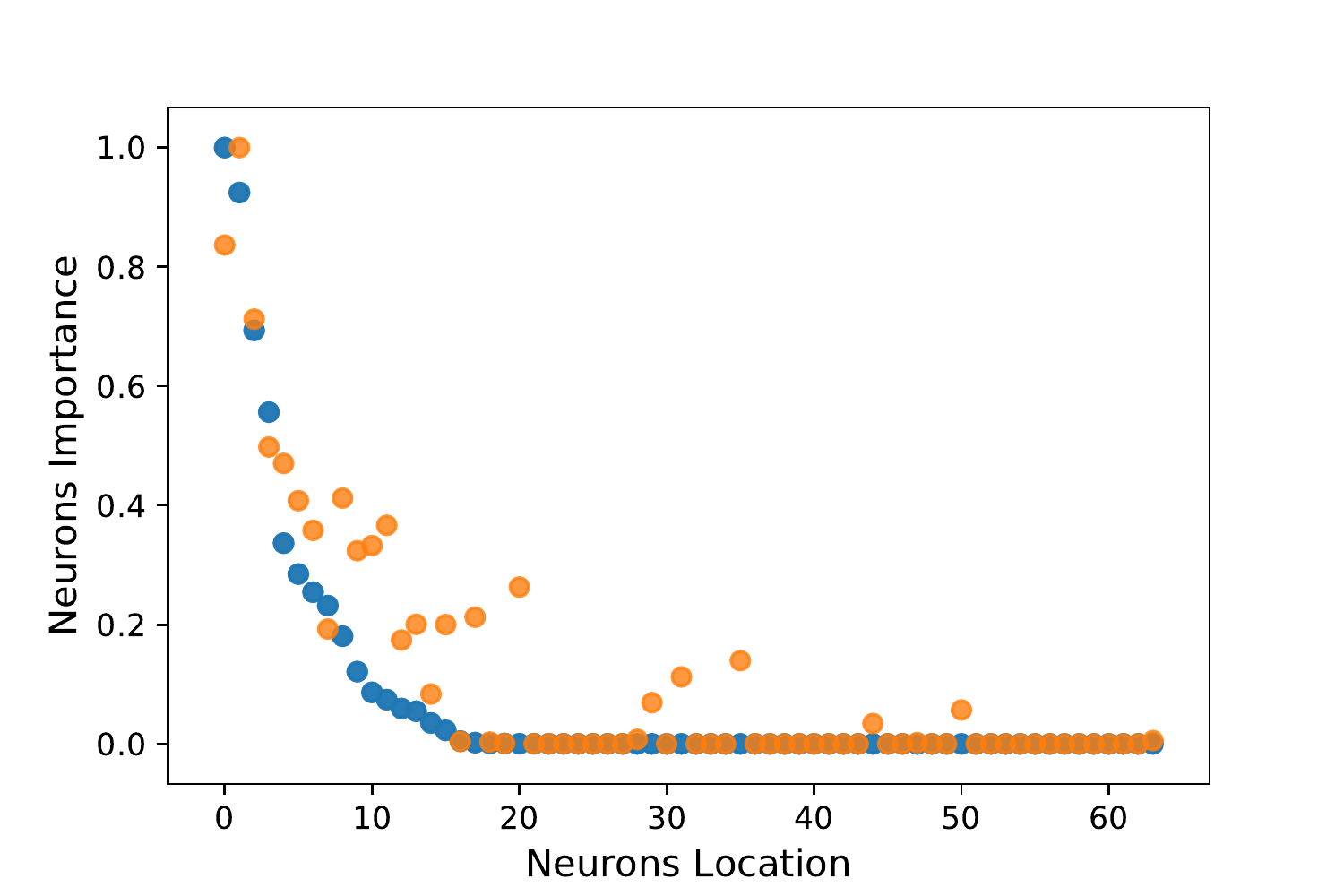} }}%
    \hfillx
    \subfloat{{\includegraphics[width=0.495\textwidth]{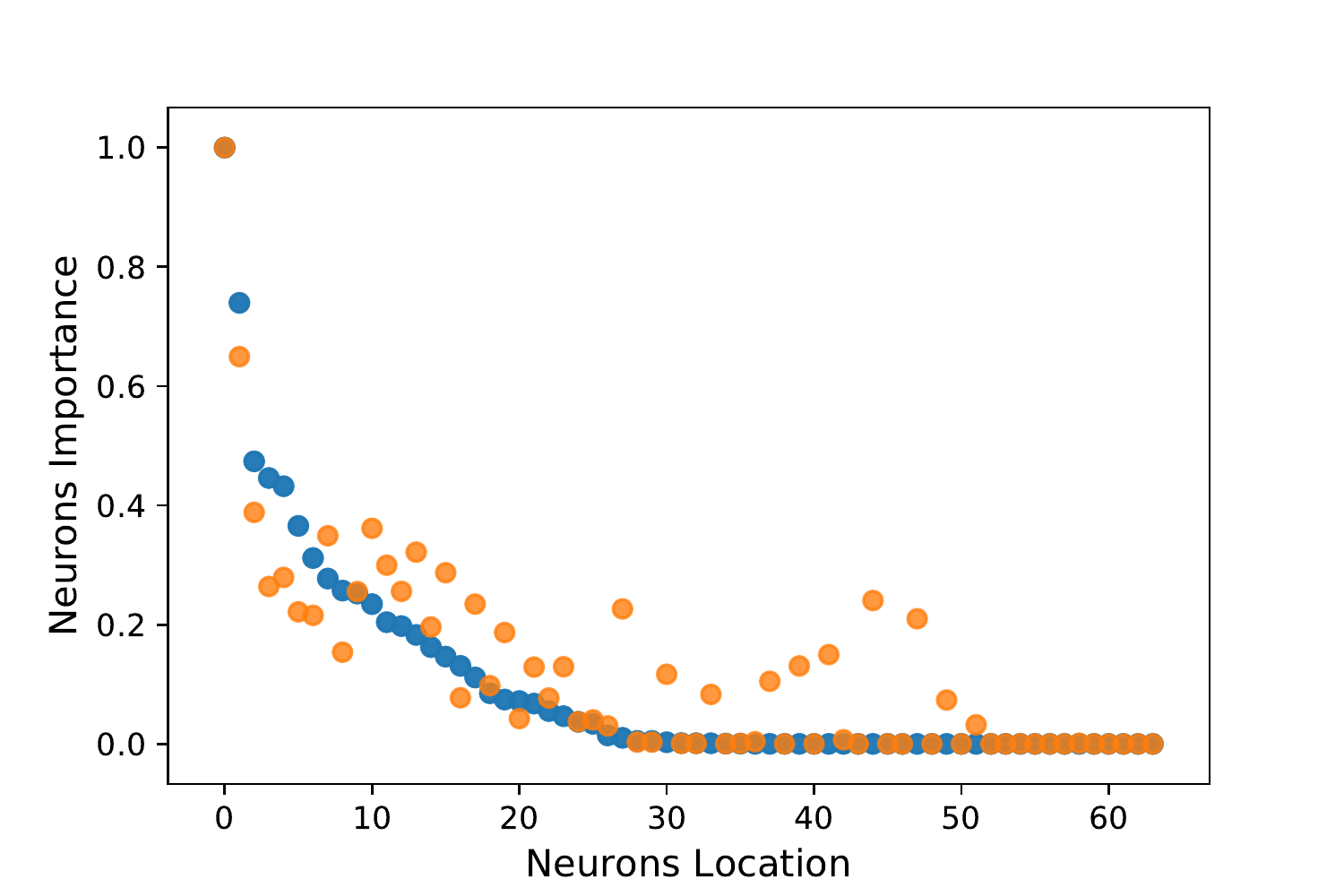} }}%
  
    \caption{\footnotesize First layer neuron importance  after learning the second task sorted in descending order according to the first task neuron importance (orange), superimposed on top of figure~\ref{fig:scale1b}. Left: \SNI, Right:  \SLNI.  \SLNI allows new neurons to become active and be used for the new task. \SNI penalizes all unimportant  neurons equally and hence more neurons are re-used then initiated for the first time. }
    \label{fig:scale2b}%
\end{figure}
\begin{figure}[h]
\vspace*{-0.4cm}
    \centering
    \subfloat{{\includegraphics[width=0.495\textwidth]{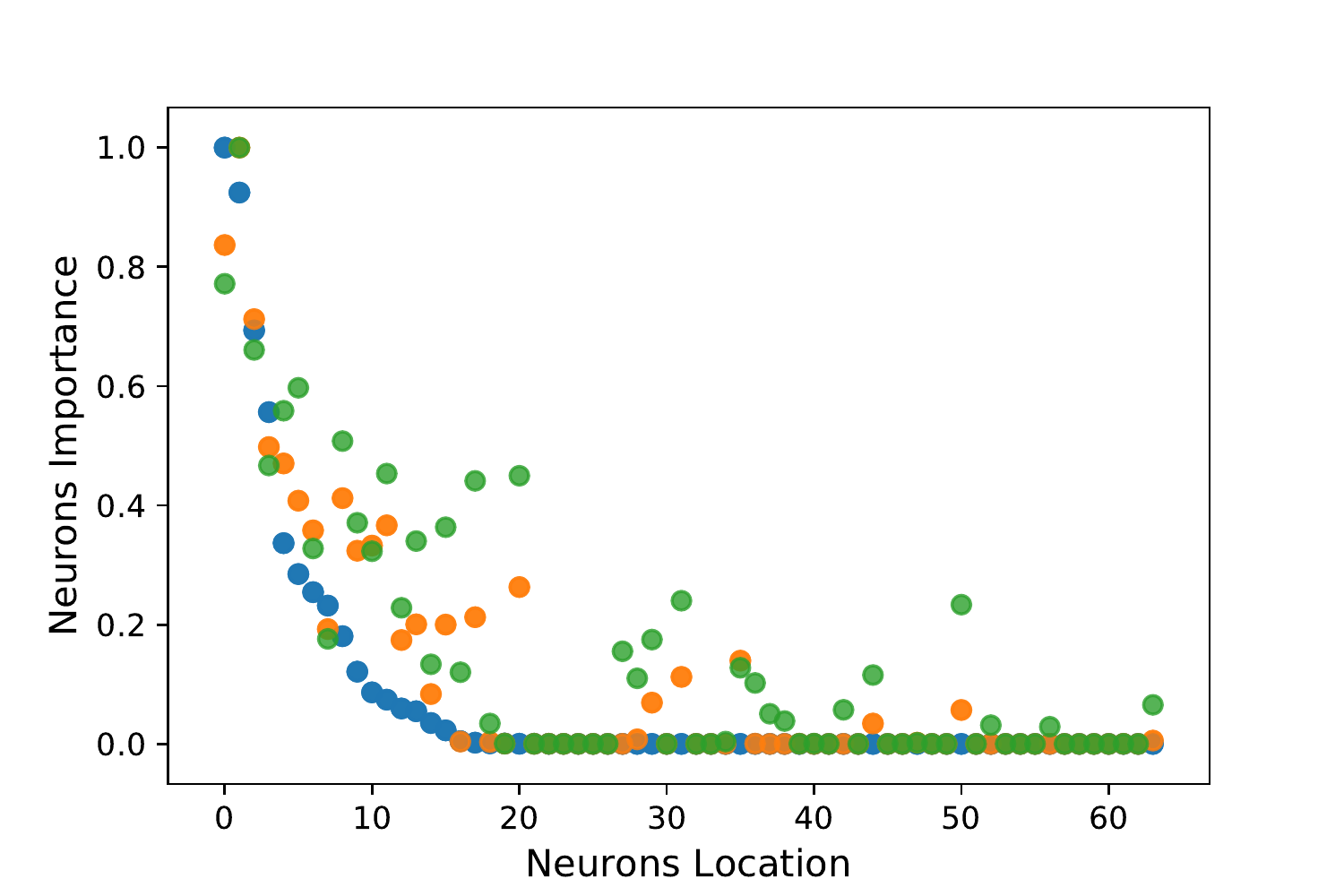} }}%
    \hfillx
    \subfloat{{\includegraphics[width=0.495\textwidth]{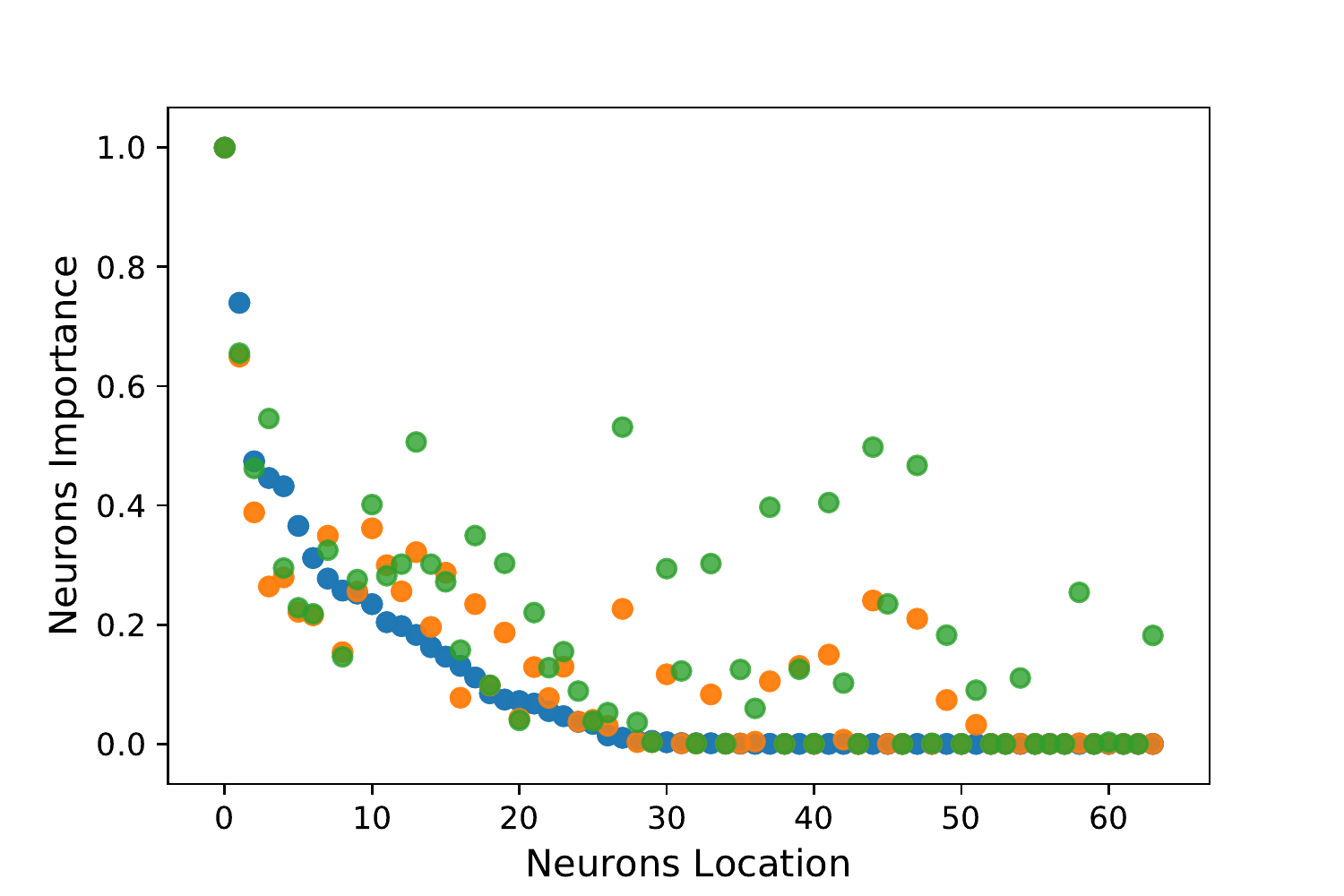} }}%
  
    \caption{\footnotesize First layer neuron importance after  learning the third task sorted in descending order according to the first task neuron importance (green), superimposed on top of figure~\ref{fig:scale2b}.  Left: \SNI, Right:  \SLNI.  \SLNI allows previous neurons to be re-used for the third task while activating new neurons to cope with the needs of the new task. For \SNI, very few neurons are newly deployed while most previous important neurons for previous tasks are re-adapted to learn the new task. }
    \label{fig:scale3b}%
\end{figure}

\end{document}